\documentclass[review,3p]{elsarticle}
\graphicspath{ {./} }
\usepackage{hyperref}
\usepackage{float}
\usepackage{verbatim} 
\usepackage{apalike}
\restylefloat{figure}
\restylefloat{table}

\usepackage{times}
\usepackage{epsfig}
\usepackage{graphicx}
\usepackage{amsmath}
\usepackage{amssymb}
\usepackage{multirow}
\usepackage{enumitem}
\usepackage{rotating}
\usepackage{array,multirow,graphicx}

\usepackage{tikz}
\tikzset{every picture/.style={font issue=\footnotesize},
         font issue/.style={execute at begin picture={#1\selectfont}}
        }
\usetikzlibrary{arrows,shapes,positioning}
\usetikzlibrary{patterns}
\usepackage{pgfplots}
\pgfplotsset{compat=1.9}
\usepackage{psfrag}
\usepackage{color, colortbl}
\definecolor{claA}{rgb}{0.95,0.95,0.95}
\definecolor{claB}{rgb}{0.90,0.90,0.90}
\definecolor{claC}{rgb}{0.80,0.80,0.80}
\definecolor{claY}{rgb}{1,1,1}
\definecolor{LightCyan}{rgb}{0.88,1,1}

\usepackage{float}
\restylefloat{table}
\usepackage{lipsum}

\journal{Expert Systems with Applications}

\biboptions{authoryear}

\begin{document}
\begin{frontmatter}

\begin{titlepage}
\begin{center}
\vspace*{1cm}

\textbf{Y-GAN: Learning Dual Data Representations for Efficient Anomaly Detection}

\vspace{1.5cm}

Marija~Ivanovska$^1$ (marija.ivanovska@fe.uni-lj.si), Vitomir~Struc$^1$ (vitomir.struc@fe.uni-lj.si) \\

\hspace{10pt}

\begin{flushleft}
\small  
$^1$ Faculty of electrical engineering, University of Ljubljana, Trzaska cesta 25, 1000 Ljubljana, Slovenia \\

\vspace{1cm}
\textbf{Corresponding Author:} \\
Vitomir~Struc \\
Faculty of electrical engineering, University of Ljubljana, Trzaska cesta 25, 1000 Ljubljana, Slovenia \\
Email: vitomir.struc@fe.uni-lj.si

\end{flushleft}        
\end{center}
\end{titlepage}

\title{Y-GAN: Learning Dual Data Representations for Efficient Anomaly Detection}

\author[label1]{Marija~Ivanovska}
\ead{marija.ivanovska@fe.uni-lj.si}

\author[label1]{Vitomir~Struc \corref{cor1}}
\ead{vitomir.struc@fe.uni-lj.si}

\cortext[cor1]{Corresponding author.}
\address[label1]{Faculty of electrical engineering, University of Ljubljana, Trzaska cesta 25, 1000 Ljubljana, Slovenia}

\begin{abstract}
We propose a novel reconstruction-based model for anomaly detection, called 'Y-GAN'. The model consists of a Y-shaped auto-encoder and represents images in two separate latent spaces. The first captures meaningful image semantics which are key for representing (normal) training data, whereas the second encodes low-level residual image characteristics. To ensure the dual representations encode mutually exclusive information, a disentanglement procedure is designed around a latent (proxy) classifier. Additionally, a novel consistency loss is proposed to prevent information leakage between the latent spaces. The model is trained in a one-class learning setting using only normal training data. Due to the separation of semantically-relevant and residual information, Y-GAN is able to derive informative data representations that allow for efficient anomaly detection across a diverse set of anomaly detection tasks.         
The model is evaluated in comprehensive experiments with several recent anomaly detection models using four popular datasets, i.e., MNIST, FMNIST, CIFAR10, and PlantVillage. 
Experimental results show that Y-GAN outperforms all tested models 
by a considerable margin and yields state-of-the-art results. The source code for the model will be made publicly available. 
\end{abstract}

\begin{keyword}
anomaly detection \sep one-class learning \sep disentangled data representations
\end{keyword}

\end{frontmatter}

\section{Introduction}
Anomaly detection represents a challenging problem, where the goal is to distinguish \textit{anomalous} data from data considered to be \textit{normal}\footnote{The term \textit{normal} describes data that conforms to some predefined characteristics and is, in general, application dependent.}~\cite{perera2021one}. Most recent solutions approach this problem from a \textit{one-class} classification perspective and attempt to learn detection models using only normal training data. Such an approach has led to successful deployment of anomaly detection techniques in a wide variety of application domains where the anomalous data is not readily available or is difficult to collect, including (visual) quality inspection \cite{racki2018compact,zavrtanik2020reconstruction}, surveillance, and security \cite{Akcay2018GANomaly,Doshi_CVPR2020_ADV_ideo,Park_CVPR2020_MemAug, sabokrou2017deep}, information forensics \cite{Khalid_CVPRW2020_OC-FakeDect, Wang_ijcai2020_FakeSpotter}, biometrics, \cite{fatemifar2019combining,oza2019active,yadav2020relativistic,perera2019learning} or medical imaging~\cite{Schlegl2017AnoGAN,Schlegl2019fAnoGAN} among others. 
\begin{figure}[!t]
\begin{center}
\centering
  \includegraphics[width=0.95\linewidth, trim = 12mm 0 0 0, clip]{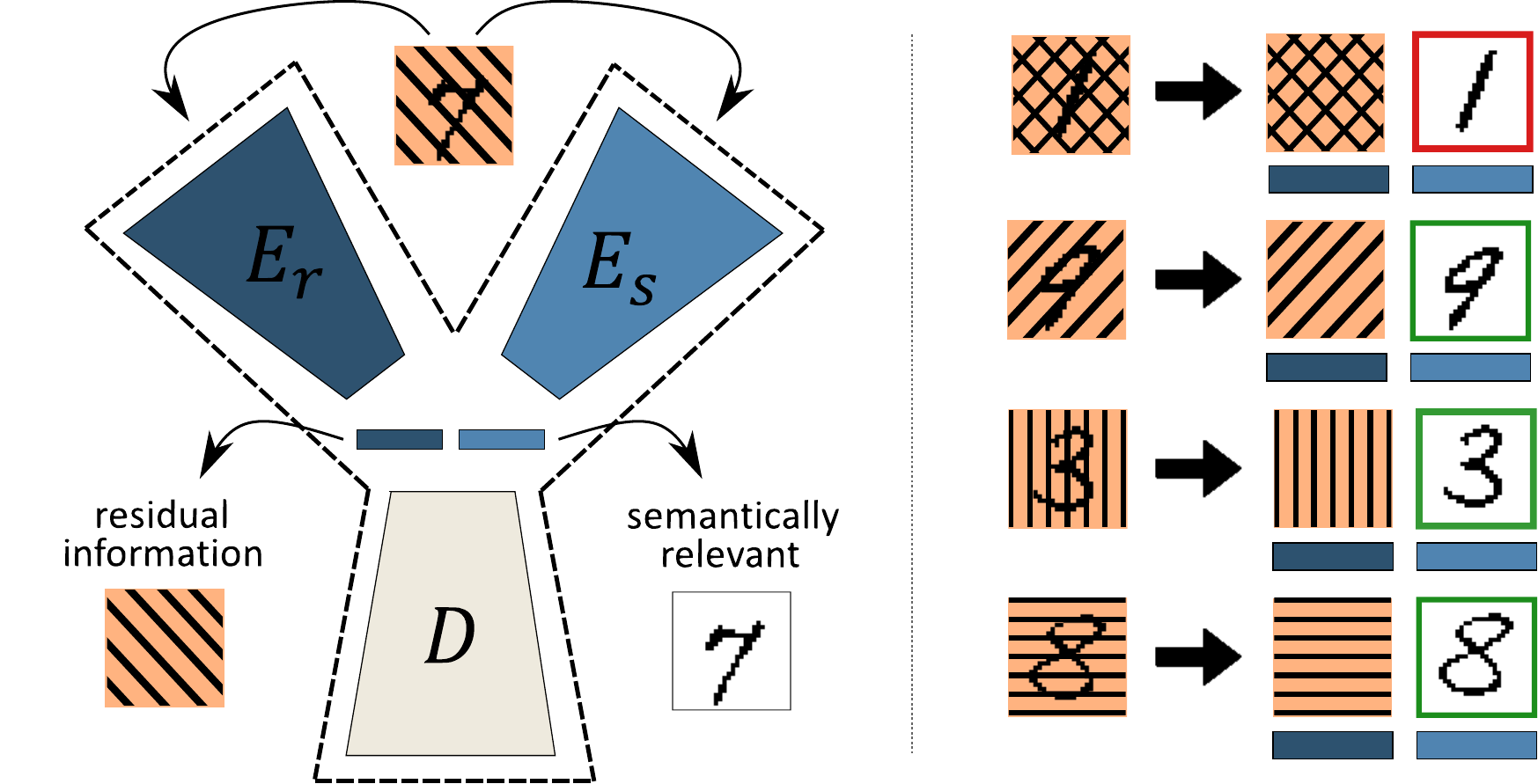}\vspace{-1mm}
\end{center}
\caption{We propose Y-GAN, a novel anomaly detection model built around a Y-shaped auto-encoder network. The model disentangles semantically-relevant image information from irrelevant, residual characteristics and facilitates efficient anomaly detection based on selective image encoding.  
As illustrated on the right, the removal of residual characteristics allows for easier detection of the digit ``$1$'', considered anomalous in this illustrative example. \label{fig:teaser} \vspace{-3mm}}
\end{figure}

Contemporary research on one-class anomaly detection is dominated by reconstruction-based  models and typically relies on powerful auto-encoders \cite{Gong_ICCV2019_MemAug_AE, Nguyen_ICCV2019_AD_AE} or generative adversarial networks (GANs) \cite{Schlegl2019fAnoGAN,P-Net_ECCV2020}. These models commonly learn some latent representation that can be used to reconstruct normal data samples with high fidelity. Because no anomalous data is seen during training, the basic assumption here is that such (anomalous) samples will lead to poor reconstructions. As a result, differences in reconstruction quality are commonly exploited to differentiate between normal and anomalous data. Reconstructive approaches have been shown to perform well across a broad range of anomaly detection tasks and to provide competitive results across several popular benchmarks~\cite{Akcay2018GANomaly,Akcay2019Skip-GANomaly}. 
However, as emphasized in~\cite{fei2020attribute}, the learning objectives typically utilized for learning reconstructive models predominantly focus on low-level pixel comparisons instead of image semantics intrinsic to the training data. This results in latent representations that encode low-level data characteristics that are likely to be shared between normal and anomalous data samples~\cite{dosovitskiy2016generating} instead of more discriminating higher-level semantics. Additionally, when data with rich visual characteristics and complex appearances is used for training, the likelihood of high-fidelity reconstructions of anomalous data increases as well, rendering reconstruction-based models less effective in such cases. This problem is further exacerbated by the high generalization capabilities of modern generative models, where high-quality reconstructions of anomalous samples can already be expected under more relaxed assumptions~\cite{fei2020attribute, Gong_ICCV2019_MemAug_AE}. The key challenge with these techniques is, therefore, to learn latent representations that encode important image semantics and are uninformative with respect to low-level visual characteristics commonly shared by normal and anomalous data.

Based on this insight, we propose in this paper a novel anomaly  detection model, called 'Y-GAN,' that aims to address the above mentioned challenges. As illustrated in Fig. \ref{fig:teaser}, Y-GAN is designed around a Y-shaped auto-encoder model that
encodes input images in two distinct latent representations. The first representation captures semantically meaningful image characteristics useful for representing key properties of normal data, while the second encodes irrelevant, residual data characteristics. This dual encoding is enabled by an efficient disentanglement procedure that can be learned automatically in a \textit{one-class learning} setting, i.e., without the use of anomalous data. To control the information content in the two latent representations, Y-GAN utilizes a latent classifier and trains it to discriminate between sub-classes/groups of normal data. In other words, it exploits differences within the normal data to learn meaningful data semantics that can later be used for anomaly detection. Additionally, a novel \textit{representation consistency} loss is introduced for the training procedure of Y-GAN that ensures that the encoded information in the dual latent representations is mutually exclusive. Using this approach, Y-GAN is able to learn highly descriptive data representations that facilitate efficient anomaly detection across a variety of problem settings.
The model is evaluated in extensive experiments on four anomaly detection benchmarks and compared with several state-of-the-art anomaly detection models presented recently in the literature. The results of the evaluations show that Y-GAN offers significant performance improvements over all considered competing models. 

In summary, our key contributions in this paper are:
\begin{itemize}
    \item We propose Y-GAN, a novel anomaly detection method, that disentangles semantically-relevant image characteristics from residual information for efficient data representation and addresses some of the key challenges associated with reconstructive anomaly-detection models. 
    \item We introduce a novel disentanglement strategy that enforces representation consistency and allows Y-GAN to exclude uninformative image information from the anomaly detection task in a \textit{one-class} learning setting. We note that the same strategy is also applicable to other problem domains in need of efficient disentanglement.
    \item We show the benefit of the proposed dual data representation over several state-of-the-art anomaly detection mechanisms by reporting superior results on multiple benchmarks and across different anomaly detection tasks. 
\end{itemize}

\section{Background and Related Work}

A considerable amount of research has been conducted in the field of anomaly detection over the years. While early appro\-aches considered statistical models \cite{eskin2000anomaly,yamanishi2004line,xu2012robust}, one-class classifiers~\cite{scholkopf2001estimating,tax2004support,lanckriet2003robust} or sparse representations \cite{cong2011sparse,lu2013abnormal,zhao2011online} for this task, more recent solutions leverage advances in deep learning to learn powerful (one-class) anomaly detectors, 
e.g., \cite{Schlegl2019fAnoGAN, abati2019latent, Ruff_2018_Deep_SVDD, Markovitz_2020_CVPR,Bergmann_2020_CVPR, Pang_2020_CVPR, Zaheer_CVPR2020_OGNet,Bergman_ICLR_2020_GOAD}. In this section, we present the most important background information with respect to 
such one-class models to provide the necessary context for our work. For a more comprehensive coverage of the area of anomaly detection and a broader discussion of existing solutions, the reader is referred to some of the excellent surveys in this field, e.g., \cite{perera2021one,chalapathy2019deep,chandola2009anomaly, pang2021deep}.  


\subsection{Reconstruction-based Anomaly Detection}
Reconstruction-based models represent one of the most widely studied groups of anomaly detectors in the 
literature. 
Such models try to discriminate between normal and anomalous data by evaluating reconstruction errors produced by generative networks trained exclusively on normal data. Schlegl \textit{et al.,}~\cite{Schlegl2017AnoGAN}, for example, proposed to project probe samples into a GAN latent space learned in this manner in their AnoGAN model and generate reconstructions from the computed latent representation for scoring. While this approach relied on two separate steps (i.e., latent-space learning and reconstruction), later improvements, such as f-AnoGAN\cite{Schlegl2019fAnoGAN} or EGBAD~\cite{Zenati2018EGBAD}, demonstrated the benefits of learning the latent representation jointly with the reconstructive mapping. Akcay \textit{et al.,}~\cite{Akcay2018GANomaly} further enhanced the capability of reconstruction-based models with an adversarial auto-encoder, called 'GANomaly.' Different from previous work, the model derived an anomaly score by comparing latent representations of original and reconstructed images to facilitate anomaly detection. Later, the same authors introduced Skip-GANomaly~\cite{Akcay2019Skip-GANomaly} (a U-Net~\cite{Ronneberger_MICCAI2015_U-Net}-based GANomaly extension) in an attempt to capture descriptive multi-scale information. \textcolor{black}{In addition to the standard reconstruction based criteria, Massoli et al.~\cite{MOCCA_TNNLS_2021} also proposed to explicitly optimize intermediate representations of each layer in the anomaly detection model.}

More recent work on reconstruction-based models capitalized on the importance of designing informative/discriminating latent spaces that can widen the gap between reconstruction errors observed with normal and anomalous data. 
Perera \textit{et al.,} \cite{Perera_2019_CVPR_OCGAN}, for example, designed a constrained latent space for their OCGAN model, such that only samples belonging to the class observed during training were reconstructed well, while anomalous samples were not. Zhao \textit{et al.}~\cite{P-Net_ECCV2020} split images into two distinct parts (i.e., texture and structure) in their  P-Net model. The two parts were then encoded 
separately with the goal of making the generated representations more informative for anomaly detection. \textcolor{black}{The model was later extended in~\cite{TNNLS_PNet_2021}, with an additional module, that memorizes the correspondence between the structure and its texture}. 
\textcolor{black}{Integrated memory modules were also utilized by Park \textit{et al.,}~\cite{Park_CVPR2020_MemAug} in their anomaly-detection approach, to lessen the representation capacity of their model for anomalous data and further improve detection results.} Conceptually similar solutions  to the work of Park \textit{et al.,} were presented in~\cite{Gong_ICCV2019_MemAug_AE, MEMGAN_Yang_2021}.




The Y-GAN model proposed in this paper, follows the general idea of reconstruction-based methods, but unlike competing solutions encodes the input data in two distinct latent representations that allow for the separation of relevant information from information irrelevant for the anomaly detection task. As we show in the following sections, this separation of information 
is: $(i)$ achieved without any assumptions regarding the source of relevant information (e.g., texture, color, structure, etc.), $(ii)$ is learned automatically in an end-to-end manner from normal data only, and $(iii)$ leads to significant performance improvements over existing reconstruction-based models on various anomaly detection tasks.       

\begin{figure*}[!t] 
\centering
  \includegraphics[width=0.85\textwidth]{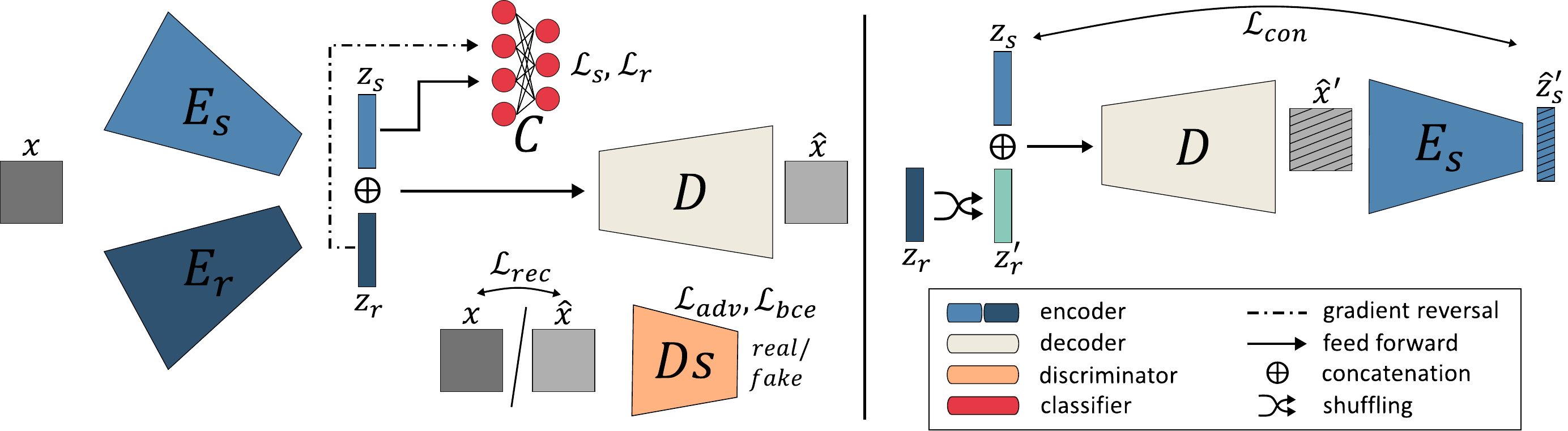}\vspace{1.5mm}
  \caption{Overview of the proposed Y-GAN model and training loss terms. The model consist of a Y-shaped auto-encoder with encoders, $E_{s}$ and $E_{r}$, for dual data representation, a decoder $D$ for image reconstruction, a latent classifier $C$ for data disentanglement, and an adversarial discriminator $Ds$, which ensures that data reconstructions $\hat{x}$ cannot be distinguished from the original input samples $x$. 
   Using an efficient disentanglement procedure, Y-GAN aims to learn a semantically meaningful data representation in $z_{s}$, that encodes only characteristics relevant for representing normal data, while capturing irrelevant, residual data characteristics in $z_{r}$. 
   The figure is best viewed in color.\label{fig:Model}\vspace{-1mm}}
\end{figure*}

\subsection{Anomaly Detection with Proxy Tasks}

To address some of the limitations of reconstruction-based anomaly detectors, another major group of existing models uses proxy tasks when learning to discriminate between normal and anomalous data.   
The main idea behind solutions from this group is that models trained on normal data will fare badly in the considered proxy task when subjected to anomalous samples. Ye \textit{et al.,}~\cite{fei2020attribute}, for instance, explored image restoration in this context and showed that differences in restoration performance can be used for anomaly detection. Noroozi \textit{et al.}~\cite{Noroozi_ECCV_2016_jigsaw} investigated jigsaw solving as a proxy task, the solutions from 
\cite{zavrtanik2020reconstruction, Haselmann_ICMLA_2018_inpainting, Stain_AD_ICPR2020, zavrtanik_ICCV2021_draem} utilized image inpainting as the proxy for anomaly detection, and the work from \cite{Bergman_ICLR_2020_GOAD, NEURIPS2019_a2b15837, GidarisSK_ICLR_2018, Golan_NIPS2018_GeoTrans} investigated classification objectives defined over self-annotated recognition problems to facilitate anomaly detection. The proposed Y-GAN is related to these models in that it also relies on a proxy classifier, which, however, aims at distinguishing between different sub-groups of the normal data. By defining the proxy task as a classification problem over normal data, Y-GAN is able to: $(i)$ ensure a \textit{compact} representation of the normal data in the model's latent space, and $(ii)$ automatically learn semantically relevant information for the anomaly detection task. Both of these characteristics are beneficial for anomaly detection performance, as we demonstrate in the experimental section.

\subsection{Pre-trained Models for Anomaly Detection}
In an effort to capture the most discriminating characteristics of the input data, many recent anomaly detection techniques try to    
leverage the representational power of features extracted by  pre-trained (large-scale) classification models. Defard \textit{et al.,}~\cite{PaDiM_ICPR2020} and Rippel \textit{et al.,}~\cite{rippel2020modeling}, for example, have shown that such features can be successfully used for explicitly modelling the distribution of normal samples. Anomalies are in this case detected as out-of-distribution samples. Similar solutions based on shallower models have also been investigated in literature, e.g., \cite{SPADE_Cohen_2020, PatchCore_Roth_2021}. Reiss \textit{et al.,}~\cite{Reiss_2021_CVPR_PANDA} proposed to fine-tune existing pretrained models using normal training data and showed that such an approach leads to impressive anomaly detection performance. Mai \textit{et al.,} \cite{mai2021brittle} explored the use of knowledge distillation to limit the set of pretrained features considered when building anomaly detection models. Similar ideas have also been pursued in 
\cite{Bergmann_2020_CVPR, wang2021student_teacher_pyramid, KD_Salehi_2020}. Wang \textit{et al.,}~\cite{Wang_2021_CVPR} extended these studies by fine-tuning the distilled student network and further improved the detection rates on several anomaly detection benchmarks. In contrast to the outlined solutions, Y-GAN does not rely on pre-trained models for data representation, but instead learns discriminating encodings from scratch by separating uninformative data characteristics from meaningful data semantics relevant with respect to the normal training data. By doing so, it is able to selectively encode part of the characteristics that are relevant for discrimination without the need for large-scale datasets and resource hungry (pre-trained) classification models.


\section{Methodology}\label{sec:methods}

The main contribution of this work is a novel (powerful) anomaly detection model, called Y-GAN. In this section we present the proposed model in detail and describe its main characteristics.   

\subsection{Proposed Model}

Y-GAN, illustrated in Fig.~\ref{fig:Model},  represents a generative adversarial network built around a Y-shaped auto-encoder. 
The key idea behind Y-GAN is to split the latent space of the auto-encoder into two distinct parts by disentangling 
informative \textit{image semantics} (e.g., shapes, appearances, textures), relevant with respect to some \textit{normal} training data from uninformative, \textit{residual image information} (e.g., noise, background, illumination changes). This separation of image content is achieved through an efficient disentanglement procedure (facilitated by a latent classifier) and allows our model to learn highly descriptive data representations for anomaly detection even in the challenging one-class learning regime.  Details on the individual components of Y-GAN are given below.

\textbf{The Auto-Encoder Network.} To be able to separate relevant image content from residual information, 
we design a Y-shaped auto-encoder network and use it as the generator for Y-GAN. As illustrated in Fig.~\ref{fig:Model}, this Y-shaped network consist of two separate encoders and a single decoder. The two encoders are identical from an architectural point of view, but have distinct parameters that can be learned independently one from the other. The first encoder $E_s$ maps the input image $x\in\mathbb{R}^{w\times h}$ into a \textit{semantically-relevant} latent representation $z_s$ and the second $E_r$ maps $x$ into the dual, \textit{residual} representation $z_r$, i.e.:
\begin{equation}
    z_{s} = E_{s}(x)\in\mathbb{R}^d \ \text{and} \ z_{r} = E_{r}(x)\in\mathbb{R}^d, 
\end{equation}
where $d$ stands for the dimensionality of $z_{s}$ and $z_{r}$. The complete latent representation of $x$ is computed as a concatenation of the two partial representations, i.e., $z = z_{s}\oplus z_{r}$, and passed to the decoder $D$ for reconstruction, i.e., 
\begin{equation}
\hat{x}=D(z)\in\mathbb{R}^{w\times h}.
\end{equation}

To ensure that all of the image content in $x$ is captured by the concatenated representation $z$, we use a standard $L_1$ reconstruction loss $\mathcal{L}_{rec}$ when learning the parameters of the auto-encoder~\cite{Isola_CVPR2017_L1_loss}:
\begin{equation} \label{eq:reconstruction_loss}
    \mathcal{L}_{rec} = \| x-\hat{x} \|_1 = \| x-D\big(E_{s}(x)\oplus E_{r}(x)\big) \|_1.
\end{equation}
Moreover, an adversarial loss term and additional learning objectives that control the information content in the latent representations $z_{s}$ and $z_{r}$ are also utilized during training. We discuss these in the following sections. 

\textbf{The Discriminator.} The expressive power of the latent representations, $z_{s}$ and $z_{r}$, critically depends on the fidelity of the image reconstructions $\hat{x}$. To improve fidelity and ensure that the reconstructed samples follow the distribution of the normal training data, we include a discriminator $Ds$ in the training procedure of Y-GAN and use an additional adversarial loss $\mathcal{L}_{adv}$ when learning the model. 
Following the recommendations from~\cite{Salimans_NIPS2016}, we update the weights of the auto-encoder, i.e., the generator of Y-GAN, based on the following feature-matching objective that reduces training instability and avoids GAN over-training, i.e.:
\begin{equation} \label{eq:adversarial_loss}
    \mathcal{L}_{adv} =  \| f(x)-f(\hat{x}) \|_2^2,
\end{equation}
where $f(\cdot)$, in our case, denotes the activations of the last convolutional layer of $Ds$. Conversely, we encourage the discriminator to distinguish between real and fake images by optimizing a standard binary cross-entropy loss: 
\begin{equation} \label{eq:bce_loss}
    \mathcal{L}_{bce} =  -\big[\log(Ds(x)) + \log(1-Ds(\hat{x}))\big].
\end{equation}

\textbf{The Latent Classifier.} 
The Y-shaped design of Y-GAN's auto-encoder allows for the partitioning of the latent space into two representations, $z_{s}$ and $z_{r}$. 
We force these representations 
to encode mutually exclusive information by using a disentanglement procedure based on a latent classifier $C$. 
The goal of this classifier is two-fold: $(i)$ to encourage semantic information relevant  for representing normal data to be encoded in $z_{s}$ and $(ii)$ to force the irrelevant residual information into the latent representation $z_{r}$.

To be able to learn $C$, we assume that the normal training data can be partitioned into $N$ sub-classes\footnote{Note that this is a reasonable assumption for many applications and holds for a wide variety of datasets and experimental protocols from the literature~\cite{perera2021one,Ruff_deepAndShallowADReview, STL-10, Abnormal101, tang2019_PCB}, including the ones used in this paper.}. The classifier is then trained to predict correct class labels from the latent representations $z_s$ and to misclassify input samples $x$ given their latent representation $z_r$. This training procedure controls the information content in the dual latent representations and helps to learn meaningful data characteristics for anomaly detection without examples of anomalous data.

A cross-entropy loss is utilized to maximize the classification performance of $C$ based on the semantically-relevant latent representation $z_{s}$, i.e.:
\begin{equation} \label{eq:classifier_loss_zs}
    \mathcal{L}_{s} =  -\sum_{i=1}^{N} y(i)\,\log(\hat{y}_s(i)),
\end{equation}
where $y\in\mathbb{R}^N$ is the one-hot 
encoded ground truth label 
of $x$ and $\hat{y}_{s}=C(E_s(x))\in\mathbb{R}^N$ is the classifier prediction.  

While minimizing $\mathcal{L}_{s}$ forces the latent representation $z_{s}$ to be informative with respect to the classification of the normal data, our goal is to achieve the opposite for $z_r$. To this end, we  transform $z_r$ with a gradient reversal layer~\cite{ganin2015_gradRev}. This transformation layer $R$ acts as an identity function in the forward pass through the model, i.e., $R(z_{r})=z_{r}$, but reverses the gradients from the subsequent layer during back-propagation, i.e., 
\begin{equation}
\frac{dR}{dz_{r}}=-\lambda_RI,
\end{equation}
where $\lambda_R$ is a hyper-parameter and $I\in \mathbb{R}^{d\times d}$ is an identity matrix. The minimization of relevant information content in $z_{r}$ is then achieved through the following objective: 
\begin{equation} \label{eq:classifier_loss_zres}
    \mathcal{L}_{r} =  -\sum_{i=1}^{N} y(i)\,\log(\hat{y}_r(i)),
\end{equation}
where $\hat{y}_{r}=C(R(E_r(x)))\in\mathbb{R}^N$.
\begin{figure}[!t]
\begin{center}
\centering
  \includegraphics[width = \columnwidth]{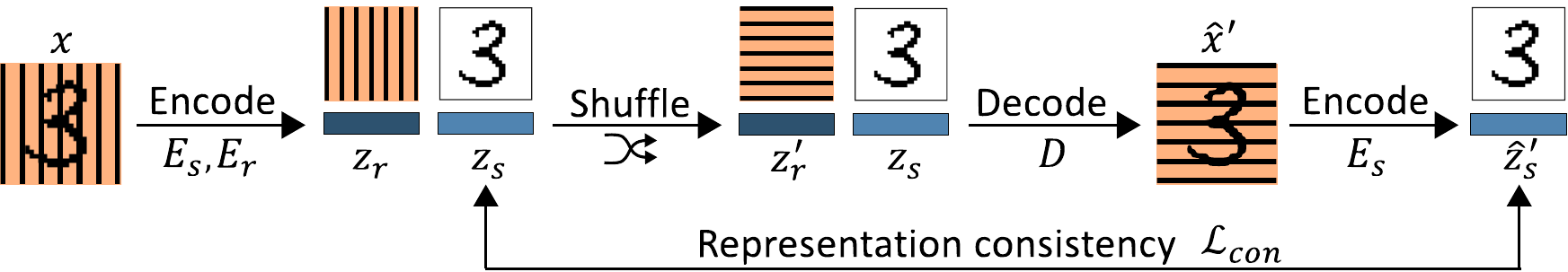}
\end{center}
\vspace{-3mm}
\caption{Illustration of the procedure used to enforce \textit{representation consistency}. The input image is represented in two latent spaces that need to encode mutually exclusive information. Y-GAN ensures that the representation in the latent representation $z_s$ is independent of that in $z_r$ by encouraging the model to produce the same  $z_s$ even when changes in $z_r$ are introduced. Shown is an illustrative toy example involving digit shapes (assumed to be relevant) and background textures (assumed to be irrelevant). \label{fig:Consistency} \vspace{-2mm}}
\end{figure}

\textbf{Enforcing Representation Consistency.} The loss functions  in Eqs.~\eqref{eq:classifier_loss_zs}~and~\eqref{eq:classifier_loss_zres} provide for a first level of disentanglement, but  do not completely prevent information leakage between the latent representations $z_{s}$ and $z_{r}$. 
For this purpose, we introduce a novel \textit{consistency loss} ($\mathcal{L}_{con}$) that penalizes the encoder $E_{s}$ in case it extracts inconsistent (semantically-relevant) information in $z_s$, when changes in the residual representation $z_r$ are introduced. The overall idea of this procedure is illustrated on the right part of Fig.~\ref{fig:Model} and using a simple visual example in Fig.~\ref{fig:Consistency}.

To calculate $\mathcal{L}_{con}$, we first randomly shuffle the set of residual representation $z_{r}$, generated from the samples in a given training batch, so that each $z_{s}$ vector is concatenated with a $z^\prime _{r}$ vector, belonging to a randomly chosen sample from the batch. These artificially created concatenations are then passed to the decoder $D$, which generates hybrid reconstructions $\hat{x}^\prime$, such that $\hat{x}^\prime = D(z_{s}\oplus z^\prime_{r})$. Next, the reconstructed images are fed to the encoder $E_{s}$, which is expected to extract latent representations $\hat{z}^\prime_{s}$ that are equivalent to the vector $z_{s}$, initially used for the generation of the hybrid reconstructions $\hat{x}^\prime$. An angular dissimilarity measure is used to penalize differences between $z_{s}$ and $\hat{z}^\prime_{s}$, 
i.e.:
\begin{equation} \label{eq:disentanglement_loss}
    \mathcal{L}_{con} =  -\frac{z_{s}\cdot \hat{z}^\prime_{s}}{\| z_{s}\| \cdot \| \hat{z}^\prime_{s} \|}.
\end{equation}
By minimizing $\mathcal{L}_{con}$, we encourage the encoder $E_{s}$ to extract image information that is invariant to the residual data characteristics encoded in $z_r$. Note that, $\mathcal{L}_{con}$ is calculated and optimized for each batch in the training set separately. The size of the training batches, therefore, has to be at least equal or greater than two, otherwise $\mathcal{L}_{con}$ has no effect. It also worth noting that $\mathcal{L}_{con}$ does not control what is encoded in the latent representations $z_{s}$ and $z_{r}$ but only ensures that the information in the two representations is mutually exclusive, or in other words, that the representations are properly disentangled. 

\subsection{Y-GAN Training}
Y-GAN is trained in an end-to-end fashion, using a multi-step procedure. For each training batch, the losses from Eqs.~(\ref{eq:reconstruction_loss}) to (\ref{eq:classifier_loss_zs}) are calculated first. Next, the set of residual representations $z_{r}$ in the given batch is randomly shuffled and processed, as described above 
for the calculation of the consistency loss $\mathcal{L}_{con}$ in Eq.~\eqref{eq:disentanglement_loss}. Finally, the weights of the generator (i.e., the Y-shaped auto-encoder) are updated, based on the combined objective $\mathcal{L_G}$, i.e.:
\begin{equation}
\mathcal{L_G}=\lambda_{1}\mathcal{L}_{rec}+\lambda_{2}\mathcal{L}_{adv}+\lambda_{3}\mathcal{L}_{s}+\lambda_{4}\mathcal{L}_{r}+\lambda_{5}\mathcal{L}_{con}.
\label{Eq:g_loss}
\end{equation}
Similarly, the weights of the adversarial discriminator $Ds$ are updated based on the combined loss $\mathcal{L_D}$, i.e.:
\begin{equation}
\mathcal{L_D}=\lambda_{1}\mathcal{L}_{rec}+\lambda_{6}\mathcal{L}_{bce}
\label{Eq:d_loss}
\end{equation}
The generator and discriminator are updated alternately for a fixed number of epochs during training.

\begin{figure}[!t]
\centering
\begin{minipage}[b]{\columnwidth}
    \centering
    \includegraphics[width=0.99\linewidth]{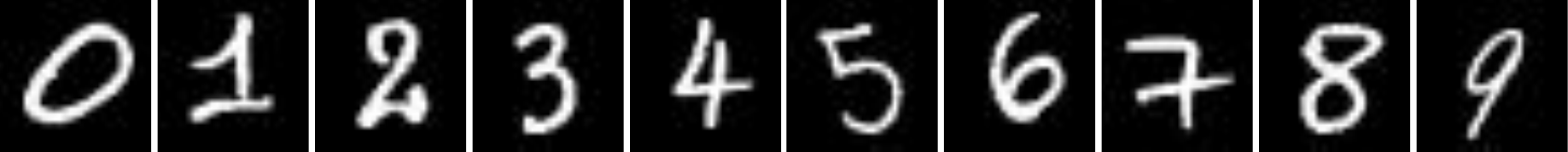}
    {\footnotesize (a) MNIST: Examples of the 10 digit categories in the dataset.}
\end{minipage}
\centering
\begin{minipage}[b]{\columnwidth}
    \centering
    \vfill \vspace{3mm}
    \includegraphics[width=0.99\linewidth]{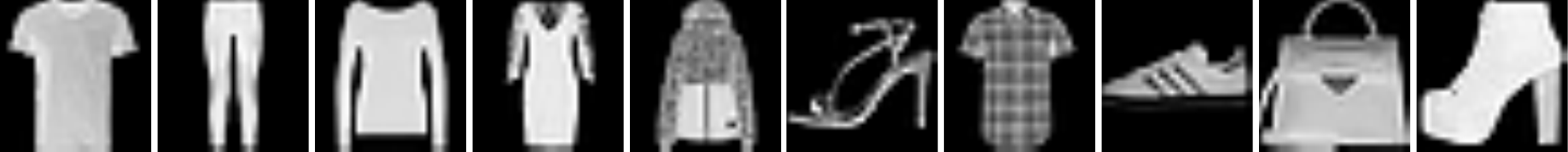}
    {\footnotesize (b) FMNIST: Examples of the 10 fashion-item categories in the dataset }
\end{minipage}
\centering
\begin{minipage}[b]{\columnwidth}
    \centering
    \vfill \vspace{3mm}
    
    \includegraphics[width=0.99\linewidth]{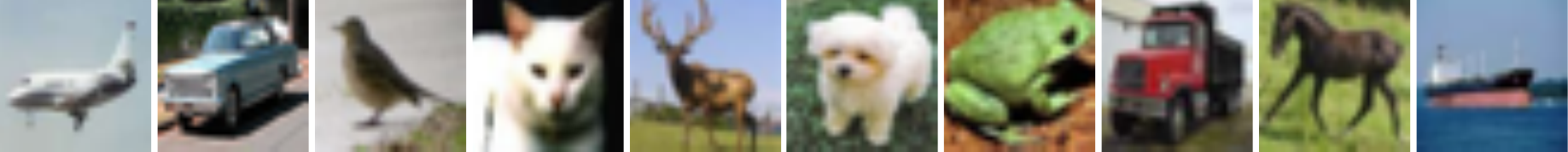}
    {\footnotesize (c) CIFAR10: Examples of the 10 object categories in the dataset }
    \vspace{-1.5mm}
\end{minipage}
\caption{Selected samples from the three standard anomaly detection benchmarks used in the experiments: (a) MNIST~\cite{MNIST_paper_lecun1998gradient, MNIST}, (b) FMNIST~\cite{FMNIST} and (c) CIFAR10~\cite{CIFAR10}. Each dataset consists of 10 different object classes.\label{fig:k_classes_out}}
\end{figure}

\subsection{Anomaly Detection with Y-GAN}\label{subsec:evaluation}
Similarly to~\cite{Golan_NIPS2018_GeoTrans}, predictions of the latent classifier $C$ are used to calculate anomaly scores. Given a probe sample $x$, the latent representation $z_{s}=E_{s}(x)$ is first computed and passed to the classifier $C$. 
Next, the activations of the output layer of the classifier are normalized $\{p_i(x)\}_{i=1}^N$, so they behave like probabilities, i.e., $\sum_{i=1}^Np_i=1$. Finally, the highest of these "probabilities" is used to compute the anomaly score $s$, i.e.:
\begin{equation} \label{eq: our_score}
s = 1- \max(p_i(x)),
\end{equation}
for the given test sample $x$. 
Due to the normalization procedure, the generated anomaly scores are bounded to $s\in[0,1]$, with $0$ representing \textit{ideal normal} data.
Note that for the calculation of the anomaly scores only $E_{s}$ and $C$ are needed, which significantly shortens inference time.

\section{Experimental Datasets and Setup}
\begin{figure}[!t]
\begin{center}
\centering
  \includegraphics[width=0.99\linewidth]{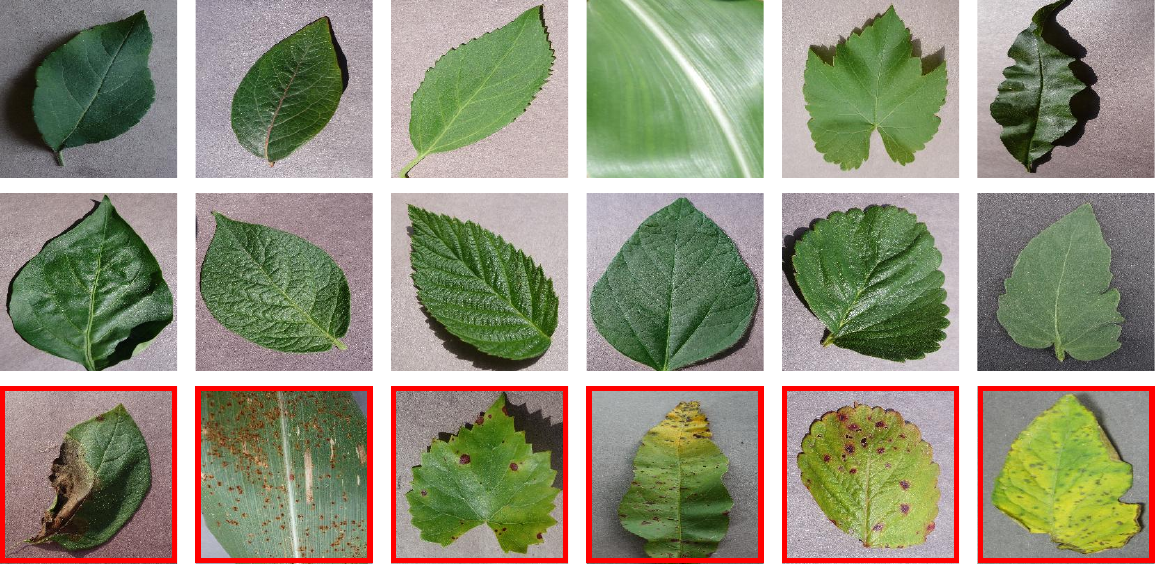}\vspace{-2mm}
\end{center}
\caption{Selected samples from the real-world dataset PlantVillage~\cite{PlantVillage_Hughes}. The dataset consists of healthy and ill leafs of $14$ different plant species. Anomalies usually represent changes in the leaf shape and color or they appear as irregular pattern on the leaf's surface. Anomalous samples are in this figure marked with red color.\label{fig:plantVillage}}\vspace{-2mm}
\end{figure}

\subsection{Datasets}
 
We evaluate Y-GAN on three standard anomaly detection benchmarks
, i.e., MNIST \cite{MNIST_paper_lecun1998gradient, MNIST}, FMNIST \cite{FMNIST}, and CIFAR10 \cite{CIFAR10}, as well as the real-world PlantVillage dataset from~\cite{PlantVillage_Hughes}. A few example images from the four datasets are presented in Figs.~\ref{fig:k_classes_out} and \ref{fig:plantVillage}. We provide details on the selected datasets below.
\begin{itemize}[leftmargin=3.5mm]
    \item \textbf{MNIST} contains $70,000$ grayscale images of handwritten digits, divided into $10$ (approximately balanced) classes, where each class represents one digit ($0$ through $9$). The images ship with a resolution of $28\times 28$ pixels and exhibit variations in terms of digit appearance \cite{MNIST_paper_lecun1998gradient, MNIST}. 
    \item \textbf{FMNIST} (Fashion MNIST) was developed as a more comprehensive alternative to MNIST. The dataset again consists of $70,000$ grayscale images of size $28 \times 28$ pixels split into $10$ balanced classes. However, FMNIST exhibits a larger degree of appearance variability than MNIST. Images in FMNIST depict clothing items grouped into different categories, i.e., T-shirts, trousers, pullovers, dresses, coats, sandals, shirts, sneakers, bags, and ankle boots \cite{FMNIST}. 
    \item \textbf{CIFAR10} contains $60,000$ color images of size $32\times32$ pixels  representing animals and vehicles from $10$ categories, i.e., airplanes, cars, birds, cats, deer, dogs, frogs, horses, ships, trucks. 
    Different from MNIST or FMNIST, images in CIFAR10 do not have a uniform background and exhibit considerable diversity in terms of appearance even within the same category. These characteristics make it particularly challenging for anomaly detection tasks \cite{CIFAR10}.
     \item \textbf{PlantVillage} is a recent real-world dataset of leafs and contains $54,305$ color images of $256\times256$ pixels. Each image represents a single leaf, photographed on a homogeneous background. Images are divided into $14$ unbalanced categories of different plant species, $9$ of which contain both, healthy and ill leafs. $3$ categories contain only healthy samples, while the remaining $2$  have no disease-free leafs. Plant diseases are usually manifested as changes in the shape and the color of the leaf, but can also appear as a subtle pattern, that covers the leaf base, as shown in 
     Fig. \ref{fig:plantVillage}.
\end{itemize}

   

The selected datasets allow for the evaluation of anomaly detection models in different problem settings, i.e.: $(i)$ with anomalies representing homogeneous classes in MNIST, FMNIST and CIFAR10, and $(ii)$ with challenging and diverse real-world anomalies in the PlantVillage dataset.
As we show in the results section, Y-GAN achieves state-of-the-art performance for both types of problems. 

\subsection{Experimental Setup}\label{SubSec: setup}

All models evaluated in the experimental section 
are trained in a {\em one-class learning} setting, where no examples of anomalous data are seen during training.
Experiments on the MNIST, FMNIST, and CIFAR10 datasets are conducted in accordance with the standard {\em k-classes-out} experimental setup
~\cite{Akcay2018GANomaly,Zenati2018EGBAD,Ruff_deepAndShallowADReview}, where nine classes are defined as normal, while the 10th class is considered anomalous. We implement the standard $80/20$ split rule commonly used in the literature~\cite{Perera_2019_CVPR_OCGAN}, where $80\%$ of the normal data is randomly selected for training, while the rest is combined with anomalous samples, forming a balanced testing set. All experiments are repeated ten times, each time with a new class defined as anomalous. Images from CIFAR10 are used with the original resolution, while MNIST and FMNIST images are re-scaled to $32\times32$ pixels to fit the Y-GAN's architecture.


For the experiments with the PlantVillage dataset, we again follow the $80/20$ split rule, by randomly selecting $80\%$ of the normal data for training purposes. The rest of the normal data along with anomalous samples is used for evaluation. However, because the number of training samples in PlantVillage is relatively small and not sufficient for deep learning tasks, the training data is augmented. For pixel-level augmentations we use techniques such as image sharpening, embossing, histogram manipulations and random changes of brightness and contrast. 
Additionally, we also apply horizontal flipping and random affine transformations. By carefully adjusting parameter values of the augmentation operations we ensure that the original dataset is significantly enlarged without inducing anomaly-like samples.   
Following prior 
work~\cite{Akcay2018GANomaly,Schlegl2017AnoGAN,  Zaheer_CVPR2020_OGNet}, we report the Area Under the Receiver Operating Characteristic (ROC) curve (AUC) as a scalar performance score in the experiments. In PlantVillage, we also report true positive (TPR) and true negative rates (TNR) for each class in the dataset. Both metrics are calculated at the equal error rate (EER) point of the ROC curve generated for all training samples from the dataset. 
\begin{figure}[!t]
\centering
\begin{minipage}[b]{0.58\columnwidth}
  \includegraphics[width=0.99\linewidth]{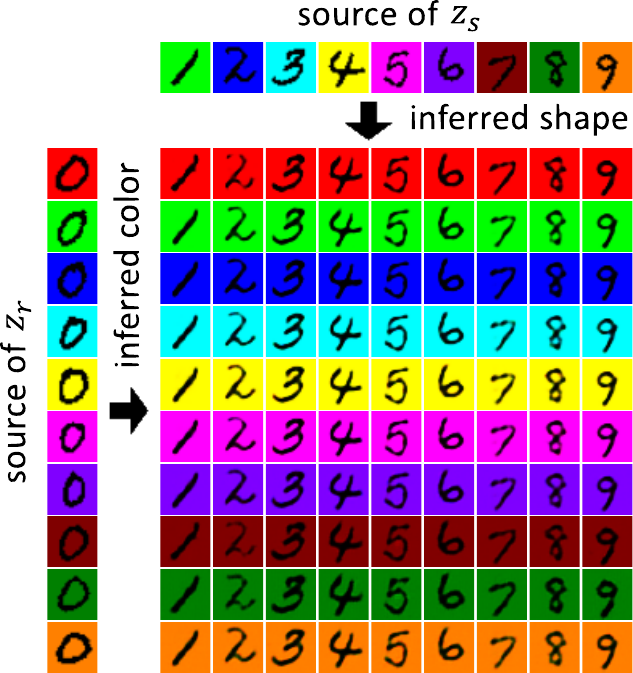}
  \centering
  {\small (a) Disentaglement analysis } 
\end{minipage}
\hfill
\hspace{0.1mm}
\begin{minipage}[b]{0.01\textwidth}
\tikz{\draw[-,gray, densely dashed, thick] (0.1,0) -- (0.1,5.7);}
\end{minipage}
\vspace{2mm}
\begin{minipage}[b]{0.36\columnwidth}
     \includegraphics[width=0.95\linewidth]{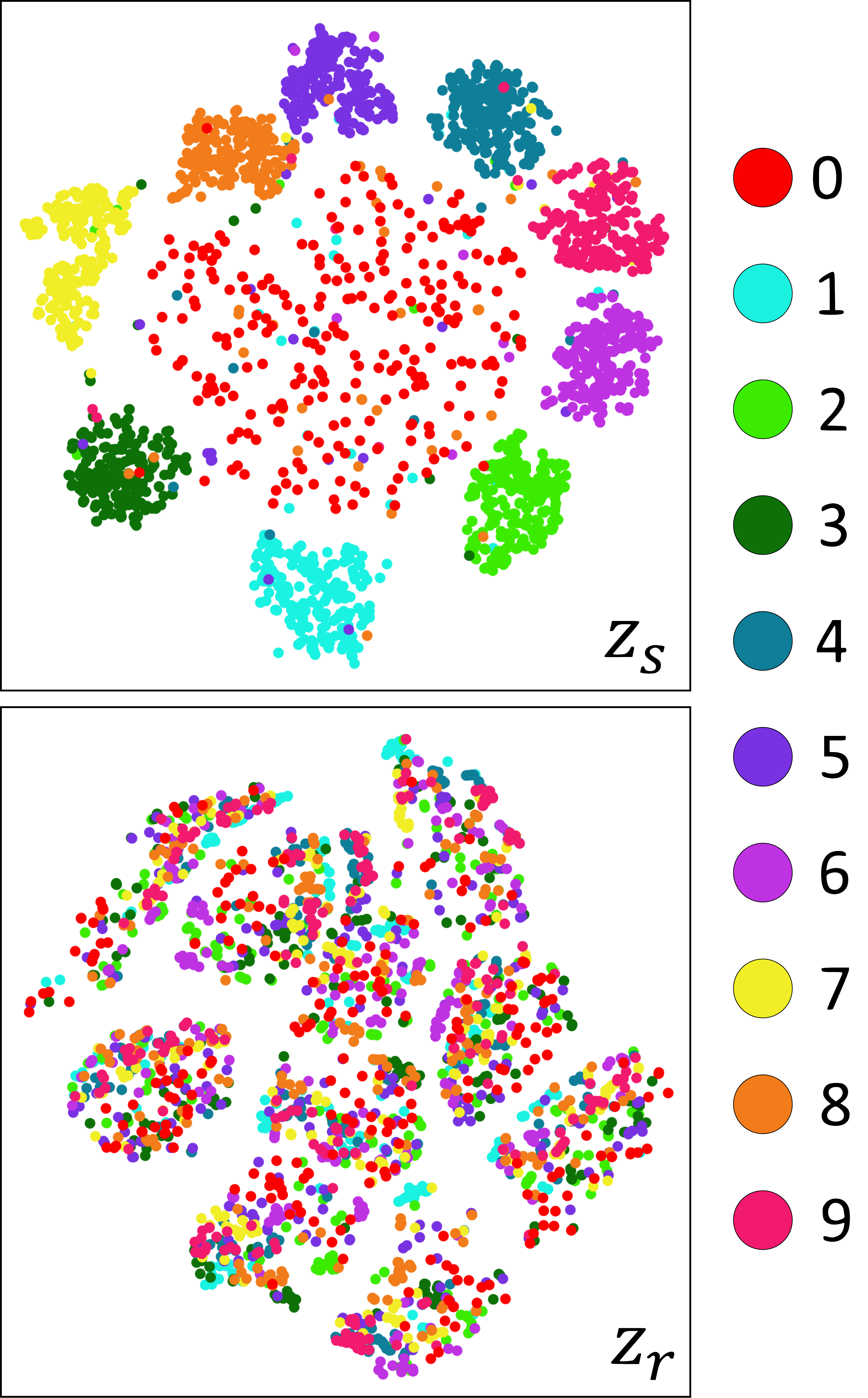}
     \centering
     {\small (b) $t$-SNE plots } 
\end{minipage}
\caption{Proof-of-concept study. Y-GAN is trained on normal data (digits $1$ to $9$) with the goal of flagging anomalous data (digit $0$). (a) Disentanglement results: the model learns to separate digits from background in the latent spaces $z_{s}$ and $z_{r}$. Shown are examples of hybrid reconstructions, where $z_{s}$ is taken from the examples on the top and $z_{r}$ from the samples on the left. (b) $t$-SNE plots in 2D: normal data forms compact well-separated clusters in the semantically-relevant latent space (marked $z_{s}$) and overlaps considerably in the residual latent space (marked $z_{r}$). }
\label{fig:disentanglement}
\end{figure}

\subsection{Implementation Details}\label{SubSec: ImplDetails}

\textbf{Model Architecture.} Y-GAN consists of five architectural building blocks, i.e., two encoders, $E_{s}$ and $E_{r}$, a decoder, $D$, a discriminator, $Ds$ and a classifier, $C$. The 
first four components are designed after 
DCGAN~\cite{Chintala_ICLR2016_DCGAN}, whereas 
the topology of the classifier is determined experimentally. 

The two encoders, $E_{s}$ and $E_{r}$, consist of convolutional layers with stride $2$. Each convolutional layer is followed by a Leaky ReLU activation with negative slope $0.2$ and a batch normalization layer. The two encoders have an identical architecture and each map the input image $x$ to a $d=100$ dimensional latent vector. 
The upscaling in the decoder $D$ is performed with transposed convolutions with stride $2$, each followed by ReLU activations and a batch normalization layer. The last convolutional layer uses an {\em tanh} activation function for bounded support.
The disciminator $Ds$ has the same architecture as $E_{s}$ and $E_{r}$, up until the last convolutional layer, which is followed by a standard \textit{sigmoid} activation.
The latent classifier $C$ is a multi-layer perceptron (MLP) with one hidden layer and $30$ hidden units. The size of the input layer is determined by the dimensionality of the latent representation, $z_{s}$, and 
the size of the output layer by the number of classes $N$ of the (normal) training data. 
In our case $N=9$ for the experiments on MNIST, FMNIST, and CIFAR10, since there are $9$ normal sub-classes defined by our experimental protocol.  For the PlantVillage $N$ is set to $N=12$, which is the number of non-anomalous plant categories in the dataset\footnote{Recall that $2$ out of the total of $14$ classes have only anomalous samples and are not considered during training.}. 


\textbf{Training Setting.} The learning objectives in Eqs. \eqref{Eq:g_loss} and \eqref{Eq:d_loss} are minimized using the Adam~\cite{Kingma_ICLR2015_Adam} optimizer with a learning rate of ${l_r}=0.0002$ and momentums $\beta_1=0.5$ and $\beta_2=0.999$~\cite{Chintala_ICLR2016_DCGAN}. The 
weights in $\mathcal{L_G}$ and $\mathcal{L_D}$ are determined empirically through an  optimization procedure on validation data. 
For the experiments we use $\lambda_{1}=\lambda_{5}=50$, and $\lambda_{2}=\lambda_{3}=\lambda_{4}=\lambda_{6}=1$.
While these weights are kept constant, the weight associated with the gradient reversal layer is initialized to a value of $\lambda_R=0$ and is then gradually increased as the training progresses, 
as suggested in~\cite{ganin2015_gradRev}. 
All models are trained for $100$ epochs on MNIST, FMNIST, and CIFAR10 and for $200$ epochs on PlantVillage, 
where less data data is available for the learning procedure. 

\textbf{Implementation.} Y-GAN is implemented in Python 3.7. using PyTorch 1.5. and CUDA 10.2. All source code, model definitions, and trained weights have been made publicly available to facilitate reproducibility\footnote{URL will be added after review.}. 
Using a personal desktop computer with an Intel\textsuperscript{\textregistered} Core\textsuperscript{TM} i7-8700K CPU and an NVIDIA\textsuperscript{\textregistered} GeForce RTX 2080 Ti GPU, it takes around four hours to train Y-GAN on MNIST, FMNIST, and CIFAR10. 
For the higher resolution images in PlantVillage, the training stage takes around five hours. Once the model is learned, a single image is processed in around $2.6\, ms$ for the smallest $32\times32$ images and $13.5\, ms$ for the largest $256\times256$ images. 

\section{Proof-of-Concept Study}

To  explore the characteristics of the dual latent-space representation and evaluate the effectiveness of the disentanglement process performed by Y-GAN, we first conduct a proof-of-concept study. To this end, we generate a color version of MNIST (\textit{Color-MNIST} hereafter) 
by inverting the thresholded black and white images of the dataset and replacing the white background 
with one of the following colors: red, green, blue, cyan, yellow, purple, violet, brown, dark green, or orange. We make sure all colors are  represented equally in the generated dataset. 
Next, we train Y-GAN on the constructed dataset by considering digits $1$ to $9$ as normal data, and $0$ as anomalous. In the test phase, we present the model with unseen samples and compute their latent representations, $z_{s}$ and $z_{r}$. Finally, we generate hybrid reconstructions by concatenating the latent vector $z_{s}$ taken from one test sample with the latent vector $z_{r}$ of another randomly selected test sample and pass the concatenated vector through the decoder. 
Example reconstructions produced with this process are shown in Fig.~\ref{fig:disentanglement}(a). 


As can be seen, Y-GAN learns to disentangle data characteristics relevant for the digit representation task, from characteristics that are irrelevant, i.e., background color in this toy example. Consequently, the replacement of the original latent vector $z_{r}$ causes a change in the background color, which is now inherited, from the randomly selected sample - shown on the left part of Fig.~\ref{fig:disentanglement}(a). Meanwhile, the shape of the digit in the original image is  preserved well. 

Next, we use $t$-distributed Stochastic Neighbor Embeddings ($t$-SNE)~\cite{van2008visualizing} to visualize the distribution of the generated data in the dual latent spaces in Fig.~\ref{fig:disentanglement}(b). Here, $250$ random samples of each of the Color-MNIST digit classes are used for visualization. Note that for the semantically-relevant latent space, samples corresponding to digits $1$ to $9$ form compact and well separated clusters (marked $z_{s}$), while samples for the anomalous $0$ are considerably less compact despite the fact that they come from a single (homogeneous) class. Nevertheless, they do not overlap (significantly) with the normal data. In the residual latent space (marked $z_{r}$),  $10$ clusters corresponding to the 
background colors used in Color-MNIST can be identified in the $t$-SNE plot. However, each cluster contains samples from all $10$ digits, suggesting that this representation has limited discriminating power for anomaly detection. 

\begin{figure}[!t]
\begin{center}
\centering
  \includegraphics[width=0.48\columnwidth,trim = 0mm 25mm 0 0, clip]{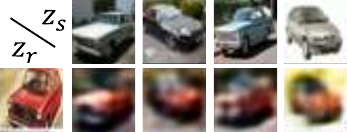}\hspace{2mm}
  \includegraphics[width=0.48\columnwidth,trim = 0mm 0 0 25mm, clip]{disentanglement_CIFAR10.pdf}\hspace{-2mm}
\end{center}
\caption{Disentanglement results on samples from two different CIFAR10 categories. The synthesized samples show that the model learns to successfully disentangle semantically relevant from residual image characteristics. Each object's shape is inferred from the source image of the semantically-relevant vector $z_s$ (first row), while color and style are inferred from the source images of $z_r$ (left most image in each example). Best viewed electronically.\label{fig:CIFAR10_disentanglement}}
\end{figure}

To test the behavior of Y-GAN on more complex data, we repeat the same experiment using CIFAR10. Selected synthesized samples and their respective source images are shown in Fig.~\ref{fig:CIFAR10_disentanglement}. We again observe that Y-GAN learned to successfully disentangle semantically relevant attributes from those that are irrelevant for representing classes in the normal data. The shape of the objects and other visually important characteristics are obviously inferred from the source image of the relevant latent vector $z_s$, while background style and colors and inferred from the residual latent vector $z_r$. Different from the Color-MNIST example, where the digits represent homogeneous classes with limited variability, the large intra-class variability of CIFAR10 images leads to lower quality reconstructions, which is expected given Y-GAN's learning objectives. Nevertheless, the results validate that meaningful separation of information content is achieved in the latent space even with challenging input images.


\section{Results and Discussion}

To  illustrate  the  performance of Y-GAN, we report in this section results that: $(i)$ compare Y-GAN to state-of-the-art techniques from the literature, $(ii)$ were generated through a comprehensive ablation study and demonstrate the contribution of various components of Y-GAN, $(iii)$ highlight some of the model's characteristics, and $(vi)$ investigate the behavior of Y-GAN in a qualitative manner.  

\begin{table*}[!t]
%
\centering
 \resizebox{0.83\textwidth}{!}{%
 \begin{tabular}{l| l | c c c c c c c c c c | c} 
\hline\hline
Model & Type$^\dagger$ & $0$ & $1$ & $2$ & $3$ & $4$ & $5$ & $6$ & $7$ & $8$ & $9$ & Mean $\pm$ Std \\  
 \hline
\rowcolor{claY} GANomaly~\cite{Akcay2018GANomaly} & RB &  $0.899$ & $0.701$ & $0.954$ & $0.820$ & $0.829$ & $0.891$ & $0.875$ & $0.735$ & $0.926$ & $0.667$ & $0.830\pm 0.094$\\
\rowcolor{claY} Skip-GANomaly~\cite{Akcay2019Skip-GANomaly} & RB& $0.845$ & $0.919$ & $0.754$ & $0.734$ & $0.530$ & $0.573$ & $0.761$ & $0.532$ & $0.765$ & $0.637$ & $0.705\pm 0.127$\\
\rowcolor{claY} OCGAN~\cite{Perera_2019_CVPR_OCGAN} & RB& {\color{red} $\mathbf{0.958}$} & {\color{red} $\mathbf{0.934}$} & {\color{red} $\mathbf{0.959}$} & {\color{red} $\mathbf{0.969}$} & {\color{red} $\mathbf{0.929}$} & {\color{red} $\mathbf{0.920}$} & {\color{red} $\mathbf{0.904}$} & $0.775$ & {\color{red} $\mathbf{0.970}$} & {\color{red} $\mathbf{0.732}$} & {\color{red} $\mathbf{0.905}\pm \mathbf{0.079} $}\\
\rowcolor{claY} f-AnoGAN~\cite{Schlegl2019fAnoGAN} & RB& $0.880$ & $0.983$ & $0.954$ & {\color{red} $\mathbf{0.969}$} & $0.928$ & $0.896$ & $0.892$ & {\color{red} $\mathbf{0.782}$} & $0.949$ & $0.702$ & $0.894\pm 0.084$\\
 P-Net~\cite{P-Net_ECCV2020} & RB & $0.788$ & $0.608$ &  $0.678$ & $0.553$ & $0.528$ & $0.467$ & $0.612$ & $0.526$ & $0.618$ & $0.474$ & $0.585\pm 0.093$\\
\rowcolor{claB} ARNet~\cite{fei2020attribute} & PT& $0.879$ & $0.798$ & $0.880$ & $0.752$ & $0.767$ & $0.816$ & $0.940$ & $0.636$ & $0.811$ & $0.685$ & $0.796\pm 0.087$\\
\rowcolor{claB} Patch SVDD~\cite{Patch_SVDD_2020_ACCV} & PT& $0.774$ & $0.709$ & $0.849$ & $0.646$ & $0.570$ & $0.656$ & $0.730$ & $0.515$ & $0.573$ & $0.609$ & $0.663\pm 0.098$\\
\rowcolor{claC} PaDiM~\cite{PaDiM_ICPR2020} & PC & $0.551$ & $0.670$ & $0.828$ & $0.647$ & $0.608$ & $0.690$ & $0.820$ & $0.779$ & $0.610$ & $0.561$ & $0.676\pm  0.097$ \\ \hline
 Y-GAN [Ours] & RB* & {\color{blue} $\mathbf{0.993}$} & {\color{blue} $\mathbf{0.993}$} & {\color{blue} $\mathbf{0.984}$} & {\color{blue} $\mathbf{0.989}$} & {\color{blue} $\mathbf{0.984}$} & {\color{blue} $\mathbf{0.986}$} & {\color{blue} $\mathbf{0.985}$} & {\color{blue} $\mathbf{0.980}$} & {\color{blue} $\mathbf{0.988}$} & {\color{blue} $\mathbf{0.987}$} & {\color{blue} $\mathbf{0.987}\pm \mathbf{0.004}$}\\
 \hline\hline
 \multicolumn{13}{l}{$^{\dagger}$\small \colorbox{claY}{RB - reconstruction based,} \colorbox{claB}{PT - proxy task-based,} \colorbox{claC}{PC - utilizing pre-trained classification models}; RB* - reconstruction-based but with a latent proxy task}
\end{tabular}
}
\caption{MNIST results in terms of AUC scores. The best model in each column is marked blue, the runner-up red.\label{tab:MNIST_results}}
\end{table*}
\begin{table*}[!t]
\centering
 \resizebox{0.88\textwidth}{!}{%
 \begin{tabular}{l | l |c c c c c c c c c c |c} 
\hline\hline
 Model & Type$^\dagger$ & T-shirt & Trousers & Pullover & Dress & Coat & Sandal & Shirt & Sneaker & Bag & Ankle boot & Mean $\pm$ Std \\  
 \hline 
\rowcolor{claY} GANomaly~\cite{Akcay2018GANomaly} & RB& $0.607$ &  {\color{red} $\mathbf{0.936}$} & $0.600$ & $0.767$ & $0.678$ & {\color{blue} $\mathbf{0.973}$} & $0.533$ & {\color{red} $\mathbf{0.895}$} & {\color{red} $\mathbf{0.961}$} & $0.867$ & $0.782\pm 0.158$\\
\rowcolor{claY} Skip-GANomaly~\cite{Akcay2019Skip-GANomaly} & RB& $0.615$ & $0.639$ & $0.698$ & $0.523$ & $0.632$ & $0.816$ & $0.612$ & $0.683$ & $0.780$ & $0.722$ & $0.672 \pm 0.082$\\
\rowcolor{claY} OCGAN~\cite{Perera_2019_CVPR_OCGAN} & RB& $0.681$ & $0.912$ & $0.643$ & $0.680$ & $0.615$ & $0.942$ & $0.563$ & $0.679$ & $0.959$ & $0.844$ & $0.752 \pm 0.140$\\
\rowcolor{claY} f-AnoGAN~\cite{Schlegl2019fAnoGAN}& RB & $0.642$ & $0.875$ & {\color{red} $\mathbf{0.758}$} & $0.656$ & $0.671$ & $0.889$ & {\color{red} $\mathbf{0.690}$} & $0.852$ & $0.945$ & {\color{red} $\mathbf{0.911}$} & $0.789\pm 0.112$\\
\rowcolor{claY} P-Net~\cite{P-Net_ECCV2020} & RB& $0.590$ & $0.586$ & $0.566$ & $0.564$ & $0.694$ & $0.340$ & $0.473$ & $0.542$ & $0.720$ & $0.714$ & $0.579 \pm 0.110$\\
 \rowcolor{claB} ARNet~\cite{fei2020attribute} & PT& $0.647$ & {\color{blue} $\mathbf{0.966}$} & $0.749$ & {\color{red} $\mathbf{0.773}$} & {\color{red} $\mathbf{0.740}$} & $0.877$ & $0.687$ & $0.751$ & $0.993$ & $0.896$ & {\color{red} $\mathbf{0.808}\pm \mathbf{0.112}$}\\
\rowcolor{claB} Patch SVDD~\cite{Patch_SVDD_2020_ACCV} & PT & \color{red} $\mathbf{0.742}$ & $0.572$ & $0.661$ & $0.597$ & $0.536$ & $0.727$ & $0.674$ & $0.577$ & $0.876$ & $0.763$ & $0.673 \pm 0.101$\\
\rowcolor{claC} PaDiM~\cite{PaDiM_ICPR2020} & PC & $0.729$ & $0.676$ & $0.475$ & $0.642$ & $0.497$ & $0.821$ & $0.502$ & $0.615$ & $0.915$ & $0.794$ & $0.667\pm 0.142$ \\ \hline
 Y-GAN [Ours] & RB* & {\color{blue} $\mathbf{0.912}$} & $0.915$ &  {\color{blue} $\mathbf{0.904}$} &  {\color{blue} $\mathbf{0.949}$} &  {\color{blue} $\mathbf{0.880}$} & {\color{red} $\mathbf{0.957}$} &  {\color{blue} $\mathbf{0.877}$} &  {\color{blue} $\mathbf{0.903}$} &  {\color{blue} $\mathbf{0.982}$} &  {\color{blue} $\mathbf{0.975}$} &  {\color{blue} $\mathbf{0.925}\pm \mathbf{0.036}$}\\
 \hline\hline
 \multicolumn{13}{l}{$^{\dagger}$\small\colorbox{claY}{RB - reconstruction-based,} \colorbox{claB}{PT - proxy task-based,} \colorbox{claC}{PC - utilizing pre-trained classification models}; RB* - reconstruction-based but with a latent proxy task}
\end{tabular}}
\caption{
FMNIST results in terms of AUC scores. The best model in each column is marked blue, the runner-up red.\label{tab: FMNIST_results}}
\end{table*}
\begin{table*}[!t]
\centering
\resizebox{0.86\textwidth}{!}{%
 \begin{tabular}{l | l |c c c c c c c c c c |c} 
\hline\hline
Model & Type$^\dagger$& Airplane & Car & Bird & Cat & Deer & Dog & Frog & Horse & Ship & Truck & Mean $\pm$ Std\\ 
\hline 
\rowcolor{claY} GANomaly~\cite{Akcay2018GANomaly} & RB& \color{red} $\mathbf{0.655}$ & $0.705$ & $0.420$ & $0.580$ & $0.359$ & $0.588$ & $0.515$ & $0.571$ & \color{red} $\mathbf{0.630}$ & $0.712$ &$0.574\pm 0.109$ \\
\rowcolor{claY} Skip-GANomaly~\cite{Akcay2019Skip-GANomaly} & RB& $0.581$ & $0.718$ & $0.494$ & $0.487$ & \color{red} $\mathbf{0.518}$ & $0.480$ & \color{red} $\mathbf{0.722}$ & $0.559$ & $0.554$ & $0.701$ & $0.581\pm 0.092$\\
\rowcolor{claY} OCGAN~\cite{Perera_2019_CVPR_OCGAN} & RB& $0.548$ & \color{red} $\mathbf{0.731}$ & $0.386$ & $0.582$ & $0.348$ & $0.597$ & $0.422$ & $0.626$ & $0.488$ & $0.713$ & $0.544\pm 0.125$\\
\rowcolor{claY} f-AnoGAN~\cite{Schlegl2019fAnoGAN} & RB& $0.578$ & $0.692$ & \color{red} $\mathbf{0.564}$ & $0.555$ & $0.459$ & $0.580$ & $0.591$ & $0.643$ & $0.610$ & $0.682$ & $0.595\pm 0.064$\\
\rowcolor{claY} P-Net~\cite{P-Net_ECCV2020} & RB& $0.582$ & $0.671$ & $0.455$ & $0.611$ & $0.476$ & $0.596$ & $0.602$ & $0.538$ & $0.523$ & $0.629$ & $0.563\pm 0.065$\\
\rowcolor{claB} ARNet~\cite{fei2020attribute} & PT & $0.598$ & $0.635$ & $0.466$ & \color{red} $\mathbf{0.706}$ & $0.435$ & \color{red} $\mathbf{0.697}$ & $0.512$ & \color{red} $\mathbf{0.662}$ & \color{red} $\mathbf{0.630}$ & \color{red} $\mathbf{0.727}$ & \color{red} $\mathbf{0.607}\pm \mathbf{0.098}$\\
\rowcolor{claB} Patch SVDD~\cite{Patch_SVDD_2020_ACCV} & PT & $0.517$ & $0.542$ & $0.505$ & $0.548$ & $0.493$ & $0.567$ & $0.529$ & $0.538$ & $0.511$ & $0.542$ & $0.529\pm 0.021$\\
\rowcolor{claC} PaDiM~\cite{PaDiM_ICPR2020} & PC & $0.542$ & $0.668$ & $0.502$ & $0.546$ & $0.335$ & $0.612$ & $0.433$ & $0.549$ & $0.386$ & $0.625$ & $0.520\pm 0.102$ \\ \hline
 Y-GAN [Ours] & RB*& \color{blue} $\mathbf{0.729}$ & \color{blue} $\mathbf{0.767}$ & \color{blue} $\mathbf{0.749}$ & \color{blue} $\mathbf{0.768}$ & \color{blue} $\mathbf{0.759}$ & \color{blue} $\mathbf{0.764}$ & \color{blue} $\mathbf{0.778}$ & \color{blue} $\mathbf{0.780}$ & \color{blue} $\mathbf{0.722}$ & \color{blue} $\mathbf{0.811}$ & \color{blue} $\mathbf{0.763\pm 0.024}$\\
 \hline\hline
  \multicolumn{13}{l}{$^{\dagger}$\small\colorbox{claY}{RB - reconstruction-based,} \colorbox{claB}{PT - proxy task-based,} \colorbox{claC}{PC - utilizing pre-trained classification models}; RB* - reconstruction-based but with a latent proxy task}
\end{tabular}}
\caption{CIFAR10 results in terms of AUC scores. The best model in each column is marked blue, the runner-up red.
\label{tab: CIFAR10_results}}
\end{table*}

\subsection{Quantitative Evaluation} 
We evaluate Y-GAN in comparative experiments with several state-of-the-art anomaly-detection models. Specifically, we compare against GANomaly~\cite{Akcay2018GANomaly}, Skip-GANomaly~\cite{Akcay2019Skip-GANomaly}, OCGAN~\cite{Perera_2019_CVPR_OCGAN}, f-AnoGAN~\cite{Schlegl2019fAnoGAN}, and P-Net~\cite{P-Net_ECCV2020,TNNLS_PNet_2021}, which represent powerful reconstruction-based (RB) anomaly detection models and are Y-GAN's \textit{main competitors}. Additionally, we also include the recent ARNet~\cite{fei2020attribute} and Patch SVDD~\cite{Patch_SVDD_2020_ACCV} approaches as representatives of proxy-task (PT) models, and the PaDiM technique from~\cite{PaDiM_ICPR2020} as an example of solutions utilizing pre-trained classification (PC) models in the evaluation. For a fair comparison, the official GitHub implementations 
are used for the experiments (where available) 
together with the advocated hyper-parameters to ensure optimal performance.

\textbf{MNIST.} The MNIST results, reported in Table \ref{tab:MNIST_results}, show that Y-GAN significantly outperforms all evaluated baselines. It improves on the mean AUC score of the runner-up OCGAN by $9\%$ and on the standard deviation (computed over all runs) by a factor of close to $20$. Compared to the competing models, Y-GAN ensures the most consistent results, regardless of which class is considered anomalous. This can be seen particularly well from the results for digits $7$ and $9$, where all tested models exhibit a drop in AUC scores, while Y-GAN retains  performance similar to other settings. 

\textbf{FMNIST.} Compared to MNIST, the AUC scores obtained on FMNIST are lower for all tested models  due to the larger image diversity in this dataset, as summarized in Table~\ref{tab: FMNIST_results}. The proposed Y-GAN achieves a mean AUC score of $0.925$, compared to $0.808$ for ARNet, $0.789$ for f-AnoGAN and  $0.782$ for GANomaly, 
which are the next three models (in this order) in terms of performance. A performance improvement of more than $14\%$ over the second best performing model, ARNet, points to the descriptiveness of the representation learnt by Y-GAN in the semantically-relevant latent space. 
Y-GAN again achieves the most consistent results across different experimental runs.       

\textbf{CIFAR10.} Images in CIFAR10 were captured in unconstrained settings, which makes 
this dataset extremely challenging, as evidenced by the results in Table~\ref{tab: CIFAR10_results}. All competing models result in mean AUC scores close (or below) to $0.6$, 
which speaks of the difficulty of learning meaningful representations on CIFAR10. The dual representation strategy of Y-GAN, on the other hand, yields a mean AUC score of $0.763$, improving on the runner-up, ARNet, by more than $25\%$. The proposed model also convincingly outperforms all competing models in all $10$ experimental runs.    

\textbf{PlantVillage.} Results for the 
PlantVillage dataset are reported in Table~\ref{tab: PlantVillage_results}. As can be seen, Y-GAN is again the top performer with an AUC of $0.962$, outperforming the state-of-the-art runner-up GANomaly by more than $23\%$. Y-GAN exceeds the detection accuracy of other models with respect to normal and anomalous samples and generates fewer misses on average as shown by the TPR and TNR scores. However, we also observe that the (global) calibration of the models results in unbalanced TPR and TNR values for certain classes (e.g., Corn, Grape, Potato). Despite these calibration issues, Y-GAN performs best on average even if the individual TPR and TNR scores are considered. We attribute this performance to the implemented dual data representation strategy, 
that allows for excluding irrelevant data characteristics when deciding whether a sample is anomalous or not.
\begin{table*}[!t]
\centering
\resizebox{\textwidth}{!}{%
 \begin{tabular}{l |l | l|c c c c c c c c c c c c c c |c} 
\hline\hline
Model & Type$^{\dagger}$ & Error & Apple & Blueberry & Cherry & Corn & Grape & Orange & Peach & Pepper & Potato & Raspberry & Soybean & Squash & Strawberry & Tomato & AUC\\ 
\hline 
\rowcolor{claY}  & & TNR & $0.228$ & $0.482$ & \color{red} $\mathbf{0.953}$ & $0.644$ & \color{blue} $\mathbf{0.941}$ & / & $0.375$ & $0.635$ & \color{blue} $\mathbf{0.968}$ & $0.853$ & \color{red} $\mathbf{0.911}$ & / & $0.750$ & \color{red} $\mathbf{0.680}$ &  \\
\rowcolor{claY} \multirow{-2}{*}{GANomaly~\cite{Akcay2018GANomaly}} & \multirow{-2}{*}{RB} &  TPR & $0.630$ & / & $0.436$ & $0.826$ & $0.864$ & \color{red} $\mathbf{0.926}$ & $0.784$ & $0.790$ & $0.601$ & / & / & $0.214$ & $0.711$ & $0.656$ & \multirow{-2}{*}{\color{red} $\mathbf{0.781}$}\\
 \hline
\rowcolor{claY}  & & TNR & \color{red} $\mathbf{0.675}$ & \color{red} $\mathbf{0.847}$ & $0.842$ & $0.253$ & \color{red} $\mathbf{0.929}$ & / & $0.222$ & \color{blue} $\mathbf{0.774}$ & $0.742$ & \color{red} $\mathbf{0.867}$ & $0.876$ & / & \color{blue} $\mathbf{0.880}$ & $0.085$ &  \\
\rowcolor{claY} \multirow{-2}{*}{Skip-GANomaly~\cite{Akcay2019Skip-GANomaly}}& \multirow{-2}{*}{RB}& TPR & $0.632$ & / & \color{red} $\mathbf{0.832}$ & \color{blue} $\mathbf{0.989}$ & $0.309$ & $0.843$ & $0.917$ & $0.578$ & $0.569$ & / & / & \color{blue} $\mathbf{0.969}$ & $0.390$ & $0.656$ & \multirow{-2}{*}{$0.746$}\\
\hline
\rowcolor{claY}  & & TNR & $0.584$ & $0.266$ & $0.877$ & $0.605$ & $0.882$ & / & $0.111$ & $0.338$ & $0.484$ & $0.640$ & $0.763$ & / & $0.315$ & $0.389$ &  \\
\rowcolor{claY} \multirow{-2}{*}{OCGAN~\cite{Perera_2019_CVPR_OCGAN}}& \multirow{-2}{*}{RB}& TPR & $0.562$ & / & $0.216$ & $0.758$ & $0.793$ & $0.426$ & $0.598$ & \color{blue} $\mathbf{0.943}$ & $0.736$ & / & / & $0.201$ & $0.886$ & $0.547$ & \multirow{-2}{*}{$0.608$}\\
\hline
\rowcolor{claY} & & TNR & $0.375$ & $0.645$ & $0.579$ & $0.224$ & $0.847$ & / & \color{red} $\mathbf{0.486}$ & $0.625$ & $0.581$ & $0.840$ & $0.767$ & / & $0.587$ & $0.260$ &  \\
\rowcolor{claY} \multirow{-2}{*}{f-AnoGAN~\cite{Schlegl2019fAnoGAN}}& \multirow{-2}{*}{RB} & TPR & $0.592$ & / & $0.465$ & $0.772$ & $0.282$ & $0.574$ & $0.655$ & $0.541$ & $0.474$ & / & / & $0.845$ & $0.192$ & $0.631$ & \multirow{-2}{*}{$0.623$}\\
\hline
\rowcolor{claY}  && TNR & $0.495$ & $0.532$ & $0.520$ & $0.528$ & $0.529$ & / & $0.444$ & $0.530$ & $0.613$ & $0.467$ & $0.526$ & / & $0.533$ & $0.489$ &  \\
\rowcolor{claY} \multirow{-2}{*}{P-Net~\cite{P-Net_ECCV2020}}& \multirow{-2}{*}{RB}& TPR & $0.508$ & / & $0.520$ & $0.517$ & $0.519$ & $0.511$ & $0.532$ & $0.524$ & $0.505$ & / & / & $0.525$ & $0.526$ & $0.517$ & \multirow{-2}{*}{$0.524$}\\
\hline
\rowcolor{claB}  && TNR & $0.608$ & $0.698$ & $0.889$ & $0.116$ & $0.824$ & / & $0.000$ & $0.767$ & \color{red} $\mathbf{0.774}$ & $0.813$ & $0.846$ & / & $0.511$ & $0.633$ &  \\
\rowcolor{claB} \multirow{-2}{*}{ARNet~\cite{fei2020attribute}}& \multirow{-2}{*}{PT}& TPR & $0.586$ & / & $0.669$ & \color{red} $\mathbf{0.979}$ & \color{red} $\mathbf{0.928}$ & $0.237$ & $0.909$ & $0.782$ & \color{red} $\mathbf{0.760}$ & / & / & $0.835$ & \color{red} $\mathbf{0.905}$ & \color{red} $\mathbf{0.672}$ & \multirow{-2}{*}{$0.736$}\\
\hline
\rowcolor{claB}  && TNR & $0.672$ & $0.299$ & $0.708$ & \color{red} $\mathbf{0.682}$ & \color{red} $\mathbf{0.929}$ & / & $0.375$ & $0.578$ & \color{red} $\mathbf{0.774}$ & $0.707$ & $0.730$ & / & $0.326$ & $0.404$ &  \\
\rowcolor{claB} \multirow{-2}{*}{Patch SVDD~\cite{Patch_SVDD_2020_ACCV}}& \multirow{-2}{*}{PT}& TPR & $0.512$ & / & $0.339$ & $0.772$ & $0.746$ & $0.512$ & $0.659$ & $0.857$ & $0.687$ & / & / & $0.335$ & \color{blue} $\mathbf{0.933}$ & $0.593$ & \multirow{-2}{*}{$0.670$}\\
\hline
\rowcolor{claC}  & & TNR & $0.605$ & $0.751$ & $0.409$ & $0.176$ & $0.529$ & / & $0.111$ & $0.507$ & $0.645$ & \color{red} $\mathbf{0.867}$ & $0.796$ & / & $0.674$ & $0.580$ &  \\ 
\rowcolor{claC} \multirow{-2}{*}{PaDiM~\cite{PaDiM_ICPR2020}}& \multirow{-2}{*}{PC}& TPR & \color{red} $\mathbf{0.736}$ & / & $0.526$ & $0.966$ & $0.878$ & $0.120$ & \color{red} $\mathbf{0.928}$ & \color{red} $\mathbf{0.887}$ & $0.754$ & / & / & $0.882$ & $0.835$ & $0.556$ & \multirow{-2}{*}{0.671}\\

 \hline
 \multirow{2}{*}{Y-GAN~[Ours]} & \multirow{2}{*}{RB*}& TNR & \color{blue} $\mathbf{0.833}$ & \color{blue} $\mathbf{0.993}$ & \color{blue} $\mathbf{0.965}$ & \color{blue} $\mathbf{1.000}$ & $0.847$ & / & \color{blue} $\mathbf{0.847}$ & \color{red} $\mathbf{0.770}$ & $0.677$ & \color{blue} $\mathbf{0.893}$ & \color{blue} $\mathbf{0.945}$ & / & \color{red} $\mathbf{0.772}$ & \color{blue} $\mathbf{0.950}$ & \multirow{2}{*}{\color{blue} $\mathbf{0.962}$} \\
 & & TPR & \color{blue} $\mathbf{0.802}$ & / & \color{blue} $\mathbf{0.909}$ & $0.800$ & \color{blue} $\mathbf{0.943}$ & \color{blue} $\mathbf{0.964}$ & \color{blue} $\mathbf{0.934}$ & $0.795$ & \color{blue} $\mathbf{0.932}$ & / & / & \color{red} $\mathbf{0.946}$ & $0.722$ & \color{blue} $\mathbf{0.926}$ &  \\
 \hline\hline
   \multicolumn{18}{l}{$^{\dagger}$\small\colorbox{claY}{RB - reconstruction-based,} \colorbox{claB}{PT - proxy task-based,} \colorbox{claC}{PC - utilizing pre-trained classification models}; RB* - reconstruction based but with a latent proxy task}
\end{tabular}}
\caption{PlantVillage results in terms of per-class TPR and TNR scores and mean AUC values over all classes. The best model in each column and for each performance score is marked blue, the runner-up is marked red. The TPR and TNR scores were computed at a (global) decision threshold defined by the equal error rate (EER) on the training data of all classes. Note that the two scores are not necessarily well calibrated for category in the dataset -- see results for Corn, Grape, or Potato leaves.
\label{tab: PlantVillage_results}}
\end{table*}

\subsection{Ablation Study}\label{SubSec: Ablation study}
To demonstrate the importance of different components of Y-GAN, 
we perform a two-part ablation study, where we first remove various parts of the learning objective $\mathcal{L_G}$, and then ablate parts of the model architecture.

\textbf{Impact of Loss Terms.} For the first part of the ablation study, three Y-GAN variants are trained with different versions of the generator loss $\mathcal{L_G}$: $(i)$ $\mathcal{L_G}$ without the consistency loss $\mathcal{L}_{con}$ (A1), $(ii)$ $\mathcal{L_G}$ without the 
residual information loss $\mathcal{L}_{r}$ (A2), and $(iii)$ $\mathcal{L_G}$ without both $\mathcal{L}_{con}$ and $\mathcal{L}_{r}$ (A3). The results in Table \ref{tab:ablation_1} show that the removal of both loss terms causes a considerable drop in the AUC scores on FMNIST, CIFAR10, and PlantVillage. 
Compared to MNIST, where a smaller drop is observed, these three datasets contain a greater amount of residual, semantically-irrelevant information (e.g., various clothing prints, background style, etc.). In such cases, both disentanglement terms, $\mathcal{L}_{con}$ and $\mathcal{L}_{r}$, play a significant role in the extraction of semantically-relevant information. Although the two losses are complementary, $\mathcal{L}_{r}$  results in a slightly greater performance drop than $\mathcal{L}_{con}$ when removed from the training objective. These results suggest that all loss terms are important and contribute to the performance of Y-GAN. 

\textbf{Impact of Architecture.} Y-GAN uses a dual encoder to generate latent representations. Four versions of the model are implemented to highlight the importance of the design choices made around this topology: 
    $(i)$ a model without the dual encoder (a single encoder $E$ is used), where the generated latent representation $z=E(x)\in\mathbb{R}^{2d}$ is split into two equally-sized vectors, $z_{s}\in\mathbb{R}^{d}$ and 
    $z_{r}\in\mathbb{R}^{d}$, on top of which $\mathcal{L_G}$~(\ref{Eq:g_loss}) is applied (B1),
    $(ii)$ the model from B1 but with a single (entangled) latent representation  - no $z_{r}$ and associated losses ($\mathcal{L}_{r}$, $\mathcal{L}_{con}$) are used (B2), 
    $(iii)$ the model from B2 but without the classifier $C$, i.e., an auto-encoder with a reconstruction-based anomaly score (B3), and 
    $(iv)$ the proposed Y-GAN model without the adversarial discriminator $Ds$, i.e., a dual encoder generator trained without $\mathcal{L}_{adv}$ (B4).
The results in Table \ref{tab:ablation_1} suggest that the removal of the dual encoder increasingly impacts results, as the complexity of the data (from an anomaly detection point of view) increases. The Y-shaped architecture, thus, contributes to a more efficient disentanglement of semantically-relevant and residual information. It can also be seen (from B2) that using a single entangled latent space representation is detrimental for the performance of the anomaly  detection  task, especially for the more challenging CIFAR10 and PlantVillage datasets. The disentanglement of irrelevant information and its removal from the decision-making process is, hence, critical for the success of Y-GAN.  The exclusion of the classifier also causes a large decrease in the overall anomaly detection accuracy across all datasets, suggesting that steered representation learning is key for Y-GAN  - see  B3 results. Finally, training the originally proposed Y-GAN auto-encoder without the adversarial discriminator, seems to have the least significant impact on the overall detection accuracy, in comparison to other Y-GAN components (B4). Nevertheless, it does contribute to the quality of the generated image reconstructions, which can further affect the performance of the disentanglement process in more complex images, e.g., in CIFAR10 and PlantVillage.  
\begin{table}[!t]
\resizebox{\columnwidth}{!}{%
 \begin{tabular}{l|cccc} 
\hline\hline
\multicolumn{1}{l|}{\textbf{Ablation study}} & MNIST & FMNIST & CIFAR10 & PlantVillage \\  
\hline
Complete Y-GAN & $0.987$ & $0.925$ & $0.763$ & $0.962 $ \\ 
\hline
\rowcolor{claB}A1: $\mathcal{L_G}$ w/o $\mathcal{L}_{con}$& $0.987$ & $0.890$ & $0.745$ & $0.918$ \\ 
\rowcolor{claB}A2: $\mathcal{L_G}$ w/o $\mathcal{L}_{r}$& $0.980$ & $0.832$ & $0.682$ & $0.865$ \\ 
\rowcolor{claB}A3: $\mathcal{L_G}$ w/o $\mathcal{L}_{r}$ and $\mathcal{L}_{con}$ & $0.962$ & $0.823$ & $0.660$ & $0.861$ \\ 
 \hline
\rowcolor{claC}B1: Y-GAN w/o dual encoders & $0.979$ & $0.881$ & $0.734$ & $0.915$ \\ 
\rowcolor{claC}B2: B1 w/o $z_{r}$& $0.956$ & $0.810$ & $0.659$ & $0.807$ \\ 
\rowcolor{claC}B3: B1 w/o $z_{r}$ and $C$ & $0.640$ & $0.689$ & $0.531$ & $0.610$ \\ 
\rowcolor{claC}B4: Y-GAN w/o $Ds$ & $0.982$ & $0.896$ & $0.719$ & $0.921$ \\ 
\hline\hline
 \multicolumn{5}{l}{Color coding: \small\colorbox{claB}{A - learning objective ablation,} \colorbox{claC}{B - architecture ablation}}
\end{tabular}}
\caption{Y-GAN ablation study. Results are reported in the form of AUC scores. The first part of the ablation study (marked A) explores the impact of loss terms, the second part (marked B) the impact of architectural components.}
\label{tab:ablation_1}
\end{table}

\subsection{Model Characteristics}

\textbf{Unlabeled Normal Data.} Y-GAN assumes that the (normal) training data comes from multiple sub-classes/groups and that labels for these sub-classes are readily available. This assumption allows for the inclusion of the latent classifier $C$ in the training process, which was shown to be critical for the overall performance of Y-GAN. 
Here, we show that it is possible to relax this assumption and train Y-GAN with unlabeled training data in a completely unsupervised manner. To this end, 
we run a clustering procedure over the training data and \textit{generate weak labels} that can be utilized when learning Y-GAN. Specifically, we first compute feature representations from the training images used in the given experimental run with the EfficientNet-B4~\cite{pmlr-EfficientNet_2019} model, pre-trained on ImageNet. Given input images $x$, the model then generates $1792$-dimensional representations  for the experiments. For efficiency reasons, we reduce the dimensionality of these representations to $100$ using Principal Component Analysis (PCA)~\cite{turk1991eigenfaces} and cluster the data with  $k$-means. We determine the optimal number of clusters based on the \textit{average silhouette method}~\cite{kaufman2009finding}. Finally, we utilize the generated cluster assignments as weak labels for  Y-GAN training.
\begin{table}[!t]
\centering
\resizebox{0.99\columnwidth}{!}{%
 \begin{tabular}{l |c c c c} 
\hline\hline
Model & MNIST & FMNIST & CIFAR10 & PlantVillage\\ 
\hline 
Y-GAN (ground truth labels) & $0.987$ & $0.925$ & $0.763$ & $0.962$\\
Y-GAN* (weak labels) & $0.964$ & $0.892$ & $0.733$ & $0.846$\\
 \hline\hline
\end{tabular}}\vspace{2mm}
\caption{Mean AUC scores generated with Y-GAN models generated with: $(i)$ ground truth labels on the sub-class structure of the normal training data, and $(ii)$ weak labels generated through a $k$-means clustering procedure. Note that even in a completely unsupervised setting, where (noisy) weak labels are inferred directly from the normal training data, Y-GAN generates state-of-the-art results that outperform all of the competing models on all four experimental datasets. 
\label{tab: results_weak}}
\end{table}
\begin{itemize}[leftmargin=3mm]
    \item \textbf{K--Classes--Out Results.}  On MNIST, FMNIST, and CIFAR10, the clustering procedure identifies either $9$ or $10$ clusters in any given experimental run ($9$ classes are in fact represented). 
An analysis of the generated clusters shows that $95\%$ of MNIST samples in each cluster share the same ground truth. This percentage is bit lower in FMNIST and CIFAR10, where it equals $91\%$ and $87\%$, respectively. 
This suggests that the clustering reasonably well approximates the actual data classes, but also that part of the data is not assigned correct class labels.
A comparison between Y-GAN trained with the ground truth labels and the weak labels generated with the clustering procedure (denoted as Y-GAN*) is presented in Table~\ref{tab: results_weak}. As can be seen, the weak labels result in slight performance degradations compared to the original Y-GAN. However, the Y-GAN* model achieves competitive results on all three datasets and still outperforms all competing baselines evaluated in Tables \ref{tab:MNIST_results}, \ref{tab: FMNIST_results} and \ref{tab: CIFAR10_results}. 
The presented results suggest that learning data representations through a latent (proxy) classifier that considers differences between different sub-classes of normal training data is beneficial for performance, even if the sub-classes are not necessarily homogeneous and contain label noise.
\item \textbf{PlantVillage Results.} For this dataset, $12$ distinct clusters are identified by $k$-means, which is corresponds to the number of original categories that include non-anomalous/normal samples. 
However, the partitioning of the PlantVillage  data is in this case less accurate in comparison to the $k$-classes-out datasets, due to similarities between objects from different classes. After the clustering process, only $80\%$ of PlantVillage images representing the same plant species and share the same ground truth label. Although such weak labels decrease the anomaly detection performance by approximately $12\%$, Y-GAN* learned without any supervision still outperforms the second best competing model, GANomaly, by $7.7\%$ (see Table~\ref{tab: PlantVillage_results} for reference). Overall, these results support the observation 
that a  competitive Y-GAN model can be trained even without access to ground truth class labels for the normal training data. 
\end{itemize}
\begin{figure}[!t] 
\centering
  \includegraphics[width=\linewidth, trim=0mm 0mm 12mm 4mm, clip]{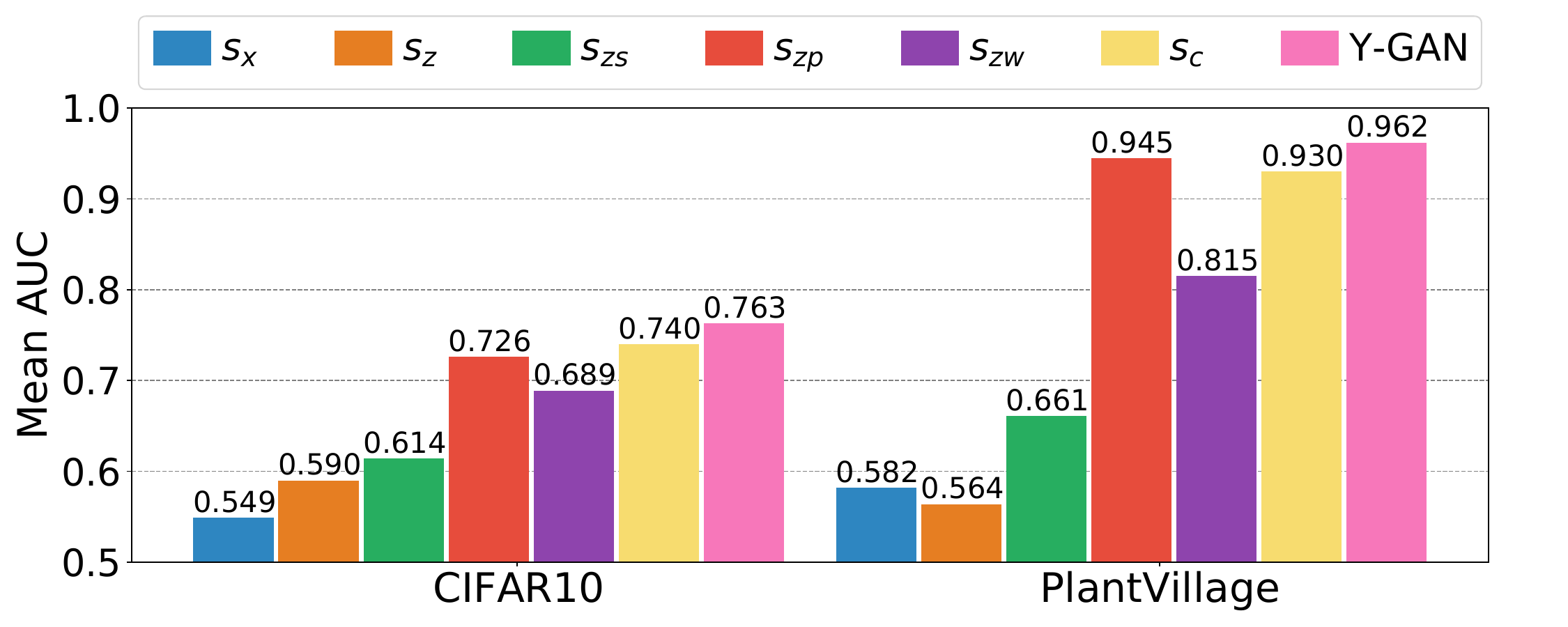}
  \vspace{-6mm}
  \caption{Score analysis on the CIFAR10 and PlantVillage datasets. Various anomaly scores, defined in the image space ($s_x$), over the latent representations ($s_z, s_{zs}$), or based on the sub-class structure of the normal data ($s_{zp},s_{zw}, s_c$, and $s$), are explored to better understand the expressive power of different representations generated within Y-GAN. Note that the proposed anomaly score achieves the strongest performance on both datasets. \label{fig: score_analysis}}
\end{figure}



\textbf{Anomaly Score Analysis.} The proposed Y-GAN uses an anomaly score derived from the output of the latent classifier $C$ to detect anomalous data. However, previous work has used other definitions of anomaly scores, including distances between the input images and their reconstructions or $L_p$ norms over latent representations among others~\cite{Perera_2019_CVPR_OCGAN,Akcay2019Skip-GANomaly}. In this section, we compare the score utilized with Y-GAN to several other possibilities. These experiment are meant to provide additional insight into the model and the characteristics of various representations generated. For the analysis, we only use the CIFAR10 and PlantVillage datasets, which contain more complex data than the other two datasets. We implement the following competing anomaly scores for the comparison:
\begin{itemize}[leftmargin=*]
    \item Image score, $s_x$: 
    \begin{equation}
        s_x(x,\hat{x}) = ||x-\hat{x}||_2^2,
    \end{equation}
    where the anomaly score is computed-based on the reconstruction quality. Here, $x$ is the input image and $\hat{x}$ is the reconstruction generated by Y-GAN.
    \item Latent score, $s_z$:
    \begin{equation}
        s_{z}(z,\hat{z}) = ||z-\hat{z}||_2^2,
    \end{equation}
    where the anomaly score is computed in the latent space using the combined latent representations $z=E_{s}(x)\oplus E_{r}(x)$ and $\hat{z}=E_{s}(\hat{x})\oplus E_{r}(\hat{x})$. $\oplus$ is a concatenation operator. 
    \item Semantic latent score, $s_{zs}$:
    \begin{equation}
        s_{zs}(z_s,\hat{z_s}) = ||z_s-\hat{z_s}||_2^2,
    \end{equation}
    where the score is computer-based on the semantic latent space representation only, i.e., $z_s = E_{s}(x)$ and $\hat{z}_s = E_{s}(\hat{x})$. 
    \item Prototype-based semantic latent score with ground truth labels, $s_{zp}$:
    \begin{equation}
        s_{zp}(z_s,z_s^{(C_i)}) = \min_i||z_s-z_s^{(C_i)}||_2^2,
    \end{equation}
    where the (semantically-meaningful) latent probe vector $z_s$ is compared to the class prototypes $z_s^{(C_i)}=1/|C_i|\sum_{z_s\in{C_i}}z_s$, computed for the $N$ sub-classes of the normal training data $\{C_i\}_{i=1}^N$ and the minimum distance is used as the anomaly score. $|\cdot|$ is the cardinality of the class.  
    \item Prototype-based semantic latent score with weak class labels, $s_{zw}$:
    \begin{equation}
        s_{zw}(z_s,z_s^{(C^*_i)}) = \min_i||z_s-z_s^{(C^*_i)}||_2^2,
    \end{equation}
    where the latent probe vector $z_s$ is compared to the class prototypes $z_s^{(C^*_i)}=1/|C^*_i|\sum_{z_s\in{C^*_i}}z_s$, computed for the $N$ sub-classes of the normal training data $\{C^*_i\}_{i=1}^N$ defined through $k$-means clustering.  The minimum distance over all class protoypes is  used as the anomaly score. 
    \item Classifier uncertainty, $s_c$:
    \begin{equation}
        s_c = -\sum_i p_i\log p_i,
    \end{equation}
    where $p=[p_1,p_2,\ldots,p_N]\in\mathbb{R}^N$ is the probability distribution for the $N$ sub-classes of the normal data computed by subjecting the output of the latent classifier $C$ to a softmax function given an input probe sample $x$.  
\end{itemize}
\begin{figure}[!t]
\centering
  \includegraphics[width=0.96\columnwidth]{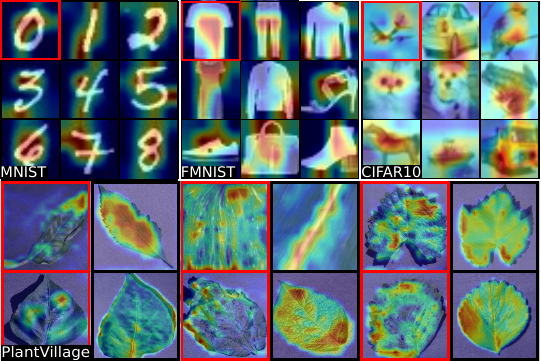}
  \caption{Sample images of correctly detected normal and anomalous samples (marked red) with corresponding Grad-CAM visualizations~\cite{Selvaraju_2017_ICCV_Grad-CAM}. As can be seen, the model focuses on 
  the global appearance of objects on MNIST and FMNIST, while local spatial characteristics are more informative on CIFAR10. On PlantVillage, on the other hand, Y-GAN appears to be simultaneously focusing on both, global and local object characteristics, due to the large intra-class variability of normal samples.\label{fig:grad_cam}}
\end{figure}

We note that all latent representations used in the above definitions of latent scores are normalized to unit norm prior to score calculation.
The results, presented in Fig.~\ref{fig: score_analysis}, show that a simple reconstruction-based score ($s_x$) results in modest performance in both datasets. The latent space score $s_z$ is slightly more informative on CIFAR10, but generates weaker results on image from PlantVillage. 
If the residual latent space is removed from the decision making process, we observe additional improvements on both datasets. Thus, the anomaly score defined in the semantically meaningful latent space $s_{zs}$  already ensures better results on CIFAR10 than all of the competing state-of-the-art models evaluated in Table~\ref{tab: CIFAR10_results} and yields comparable detection results as a large portion of the tested models on PlantVillage. If anomaly scores are defined by also considering the sub--classes present in the (normal) training data (i.e., $s_{zp}$, $s_{zw}$, $s_{c}$, and the proposed Y-GAN score $s$), we see another significant jump in AUC results on both datasets, which suggest that the structure (or distribution) of the normal data is an important source of information that can be exploited to improve anomaly detection performance. 
\begin{figure}[!t]
\begin{center}
    \includegraphics[width=0.96\columnwidth]{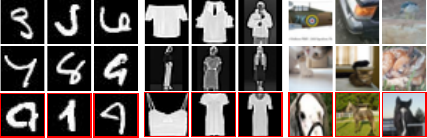}\vspace{-2mm}
\end{center}
  {\scriptsize \hspace{3.5mm} (a) Anomaly: Digit $4$ \hspace{5mm} (b) Anomaly: Pullover \hspace{6mm} (c) Anomaly: Horse}\vspace{-1mm}
  \caption{Examples of edge cases with the $k$-classes-out datasets (MNIST, FMNIST, CIFAR10). Undetected normal samples in the top two rows exhibit visual similarities with the anomalous class or are poor representatives of normal data. Similarly, the appearance of undetected anomalous samples in the bottom row (red) is close to the appearance of classes in the normal data. \label{fig:failure_cases_standard_data}}
  \vspace{-5mm}
\end{figure}

 
\subsection{Visual/Qualitative Evaluation}
\textbf{Grad-CAM Analysis.} We conduct a qualitative analysis to gain better insight into the behavior of the proposed Y-GAN model. To this end, we generate Grad-CAM visualizations~\cite{Selvaraju_2017_ICCV_Grad-CAM} of image regions that are most informative with respect to the anomaly detection task. 
This can be done due to the use of the latent (proxy) classifier $C$ utilized for the computation of anomaly scores. Examples of correctly classified normal and anomalous samples (marked red) with superimposed heatmaps are shown in Fig.~\ref{fig:grad_cam}. As can be seen, the global appearance of the objects is critical for anomaly detection on datasets with homogeneous backgrounds and small intra-class variability, such as MNIST and FMNIST. Conversely, detection on datasets with more complex visual data (such as CIFAR10) is primarily based on local object and texture characteristics. Different from MNIST, FMNIST, and CIFAR10, PlantVillage exhibits relatively large intra-class variability in terms of the size, shape, illumination, and orientation of the leaves in the images. Additionally, anomalies can  appear either as inconsistencies in the overall shape and color or, alternatively, impact the texture of the leafs at an arbitrary spatial location in this dataset. Therefore, both global and local image characteristics appear to play an important role in the detection of anomalous leaves, as seen in Fig.~\ref{fig:grad_cam}. Interestingly, Y-GAN is able to adapt to the anomaly detection task and learn descriptive and informative features from the input data regardless of whether these features correspond to global or local (or both) image characteristics.    
\begin{figure}[!t]
\centering
  \includegraphics[width=0.9\columnwidth]{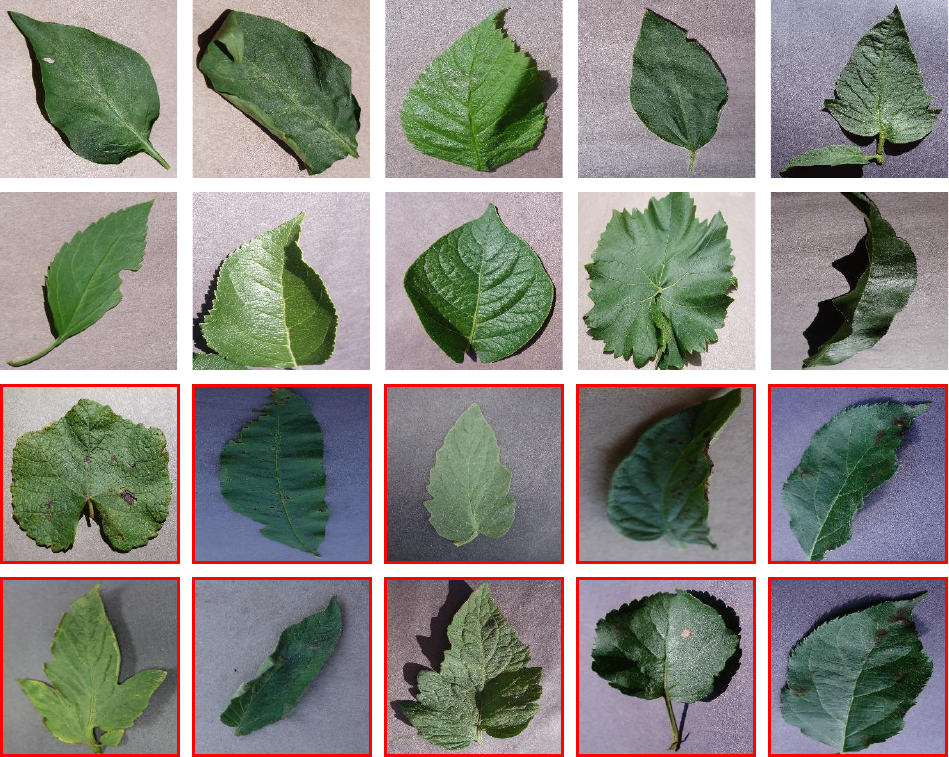}
  \caption{Sample images of undetected normal and anomalous samples (marked red) from the PlantVillage dataset. Errors with the normal data occur due to uncommon leaf shapes, holes in healthy leafs, and unusual positions/orientations. With anomalous samples, problems appear due to subtle, often unnoticeable changes in leaf color or local textures.  \label{fig:failure_cases}}
\end{figure}

\textbf{Visual Evaluation.} To better understand why the model fails to classify certain normal and anomalous samples, we perform an additional visual inspection of a few edge cases from the four experimental datasets. in Figs. \ref{fig:failure_cases_standard_data} we show results for the $k$-classes out datasets MNIST, FMNIST, and CIFAR10, where the class listed below the images was considered anomalous with the presented examples. As can be seen, the undetected normal samples correspond to objects with uncommon appearance for the considered normal data, i.e., oddly-shaped digits for MNIST, ambiguous fashion classes for FMNIST, and unusual object appearancs for CIFAR10. Difficult anomalous samples, conversely, often resemble certain classes from the normal data or exhibit ambiguous appearance. Fig. \ref{fig:failure_cases} present edge cases for the PlanVillage dataset. Here, severely folded healthy leaves and distorted leaf shapes are often detected as anomalous. Similar outcomes are also observed with leaves with holes, although such holes do not necessarily indicate an illness. Shadows darkening various parts of non-anomalous leaves can also trick the model into misclassifying normal samples. Conversely, undetected anomalies typically represent subtle, unnoticeable changes in the leaf color or local textures. 

\section{Conclusion}

The paper introduced a reconstruction-oriented auto-encoder based anomaly detection model, called Y-GAN. Different from competing approaches, Y-GAN learns to disentangle image characteristics that are relevant for representing normal data from irrelevant residual data characteristics and derives anomaly scores from selectively encoded image information. The model was shown to significantly outperform several state-of-the-art anomaly detection models on the MNIST, FMNIST, CIFAR10 and  PlantVillage 
datasets and provide the most consistent performance across different anomaly detection tasks among all tested models. As part of our future work, we plan to extend the model, so it allows for additional functionality, such as anomaly localization/segmentation, which is of interest for various anomaly detection tasks.


\section*{Acknowledgements}
Supported in parts by the ARRS Research Program P2-0250 (B) 
and  the ARRS Research Project J2-9433 (B).

\bibliography{template.bib}{}

\end{document}